\documentclass[12pt]{article}
\usepackage[letterpaper,top=2cm,bottom=2cm,left=3cm,right=3cm,marginparwidth=1.75cm]{geometry}

\usepackage{amsmath,amsfonts,bm}

\def\eqref#1{equation~\ref{#1}}

\def\1{\bm{1}}

\DeclareMathAlphabet{\mathsfit}{\encodingdefault}{\sfdefault}{m}{sl}
\SetMathAlphabet{\mathsfit}{bold}{\encodingdefault}{\sfdefault}{bx}{n}

\newcommand{\R}{\mathbb{R}}

\DeclareMathOperator*{\argmax}{arg\,max}
\DeclareMathOperator*{\argmin}{arg\,min}

\usepackage[english]{babel}
\usepackage[utf8]{inputenc} 
\usepackage[T1]{fontenc}    
\usepackage{hyperref}       
\usepackage{url}            
\usepackage{xurl}
\usepackage{booktabs}       
\usepackage{nicefrac}       
\usepackage{microtype}      
\usepackage{amsthm,amsmath,amsfonts,amssymb}
\usepackage{graphicx}
\usepackage{xcolor}
\usepackage{mathrsfs,mathtools}
\usepackage{bm}
\usepackage{adjustbox}

\usepackage{comment}
\usepackage{threeparttable}
\usepackage{bbm}
\usepackage{authblk}
\usepackage{natbib}

\usepackage{caption}
\usepackage{subcaption}

\usepackage{algorithm}
\usepackage{algorithmic}

\newlength\myindent
\newcommand{\rebut}[1]{\textcolor{black}{#1}}
\newcommand{\rebutkdd}[1]{\textcolor{black}{#1}}

\newcommand*{\eg}{\emph{e.g.}{}}
\newcommand*{\ie}{\emph{i.e.}{}}

\newtheorem{prop}{Proposition}
\newtheorem{lemma}{Lemma}

\newtheorem{rmk}{Remark}
\newcommand{\AB}{\overline{A}}
\newcommand{\XB}{\overline{X}}
\newcommand{\xb}{\overline{x}}
\newcommand{\ab}{\overline{a}}
\newcommand{\EE}{\mathop{\mathbb{E}}}
\newcommand{\prob}[1]{f\left(#1\right)}
\newcommand{\cprob}[2]{f_{#1}\left(#2\right)}
\newcommand{\indep}{\perp \!\!\! \perp}

\title{\LARGE \bf Counterfactual Generative Models\\for Time-Varying Treatments}

\author[1]{Shenghao Wu}
\author[1]{Wenbin Zhou}
\author[2]{Minshuo Chen}
\author[1]{Shixiang Zhu}
\affil[1]{
Carnegie Mellon University}
\affil[2]{
Princeton University
}

\allowdisplaybreaks

\begin{document}
\maketitle

\begin{abstract}
Estimating the counterfactual outcome of treatment is essential for decision-making in public health and clinical science, among others. Often, treatments are administered in a sequential, time-varying manner, leading to an exponentially increased number of possible counterfactual outcomes.
Furthermore, in modern applications, the outcomes are high-dimensional and conventional average treatment effect estimation fails to capture disparities in individuals. 
To tackle these challenges, we propose a novel conditional generative framework capable of producing counterfactual samples under time-varying treatment, without the need for explicit density estimation. 
Our method carefully addresses the distribution mismatch between the observed and counterfactual distributions via a loss function based on inverse probability re-weighting, and supports integration with state-of-the-art conditional generative models such as the guided diffusion and conditional variational autoencoder.
We present a thorough evaluation of our method using both synthetic and real-world data. 
Our results demonstrate that our method is capable of generating high-quality counterfactual samples and outperforms the state-of-the-art baselines.
\end{abstract}

\section{Introduction}

\begin{figure*}
\centering
\includegraphics[width=\linewidth]{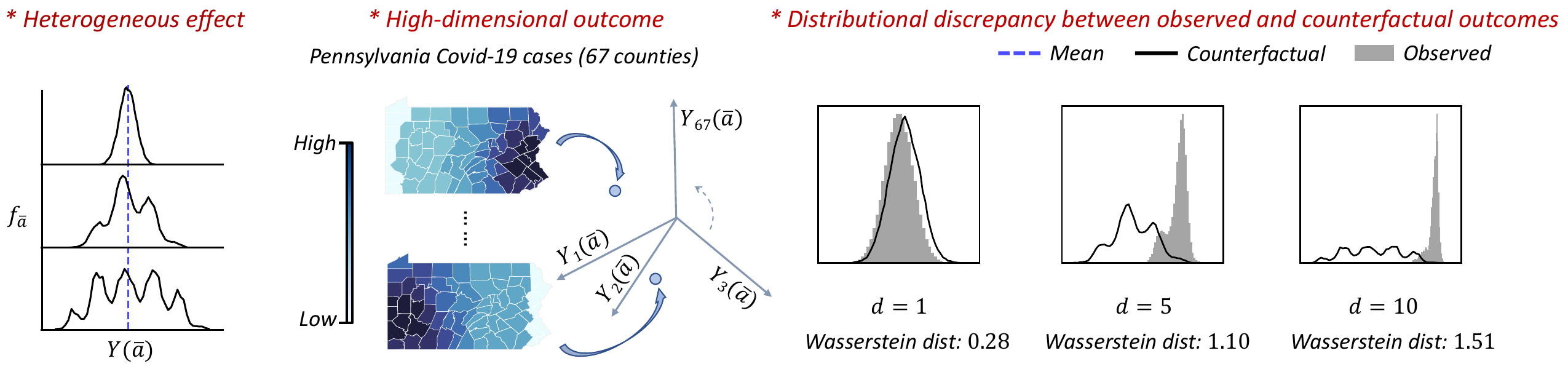}
\caption{Challenges in estimating the counterfactual outcomes of time-varying treatments. \textit{Left}: The mean is incapable of describing the heterogeneous effect in counterfactual distributions. \textit{Middle}: In a realistic scenario where the treatment is the state-level mask mandate, the outcome is a $67$-dimensional vector, corresponding to the number of COVID-19 cases of the $67$ counties in Pennsylvania. \textit{Right}: The longer the dependence on the treatment history, the greater the distributional mismatch tends to be. Here $d$ denotes the length of the history dependence. }
\label{fig:illusration}
\end{figure*}

Estimating the time-varying treatment effect from observational data has garnered significant attention due to the growing prevalence of time-series records. 
One particular relevant field is public health \citep{kleinberg2011review, zhang2017mining, bonvini2021causal}, where researchers and policymakers grapple with a series of decisions on preemptive measures to control epidemic outbreaks, ranging from mask mandates to shutdowns.
It is vital to provide accurate and comprehensive outcome estimates under such diverse time-varying treatments, so that policymakers and researchers can accumulate sufficient knowledge and make well-informed decisions with discretion.

In the literature, average treatment effect estimation has received extensive attention and various methods have been proposed \citep{rosenbaum1983central,hirano2003efficient, imbens2004nonparametric, lim2018forecasting,bica2019estimating, berrevoets2021disentangled, seedat2022continuous, melnychuk2022causal, frauen2023estimating, vanderschueren2023accounting}. By estimating the average outcome over a population that receives a treatment or policy of interest, these methods evaluate the effectiveness of the treatment via hypothesis testing. However, solely relying on the average treatment effect might not capture the full picture, as it may overlook pronounced disparities in the individual outcomes of the population, especially when the counterfactual distribution is heterogeneous (Figure~\ref{fig:illusration}, left).

Recent efforts \citep{kim2018causal, Kennedy23, melnychuk2023normalizing} have been made to directly estimate the counterfactual density function of the outcome. 
This idea has demonstrated appealing performance for univariate outcomes. Nonetheless, for multi-dimensional outcomes, the estimation accuracy quickly degrades
\citep{scott1983probability}. 
In modern high-dimensional applications, for example, predicting COVID-19 cases at the county level of a state (Figure~\ref{fig:illusration}, middle), these methods are hardly scalable and incur a computational overhead.

Adding another layer of complexity, considering time-varying treatments causes the capacity of the potential treatment sequences to expand exponentially. 
For example, even if the treatment is binary at a single time step, the total number of different combinations on a time-varying treatment increases as $2^d$ with $d$ being the length of history. More importantly, time-varying treatments lead to significant distributional discrepancy between the observed and counterfactual outcomes (Figure~\ref{fig:illusration}, right). 

In this paper, we provide a whole package of accurately estimating high-dimensional counterfactual distributions for time-varying treatments. Instead of a direct density estimation, we implicitly learn the counterfactual distribution by training a generative model, capable of generating credible samples of the counterfactual outcomes given a time-varying treatment. This allows policymakers to assess a policy's efficacy by exploring a range of probable outcomes and deepening their understanding of its counterfactual result. Here, we summarize the benefits of our proposed method:
\begin{enumerate}
    \item Our model is capable of handling high-dimensional outcomes, surpassing existing top-performing baselines in estimation accuracy and counterfactual sample quality.
    
    \item Our model's generative capability uncovers the multi-modality of the high-dimensional counterfactual samples, such as identifying distinct disease outbreak hotspots across U.S. counties.

    \item Applying our model to real COVID-19 mask mandate data shows that full mask mandates lead to significantly higher variance than not having a mandate, highlighting our method's strong potential in policy-making uncertainty quantification.
\end{enumerate}




To be specific, we develop a conditional generator \citep{mirza2014conditional, sohn2015learning, ho2022classifier}. This generator, which we choose in a flexible manner, takes into account the treatment history as input and generates counterfactual outcomes that align with the underlying distribution of counterfactuals.
The key idea behind the scenes is to utilize a ``proxy'' conditional distribution as an approximation of the true counterfactual distribution. 
To achieve this, we establish a statistical relationship between the observed and counterfactual distributions inspired by the g-formula \citep{neyman1923applications, rubin1978bayesian, robins1999association,fitzmaurice2008longitudinal}. 
We learn the conditional generator by optimizing a novel weighted loss function based on a pseudo population through Inverse Probability of Treatment Weighting (IPTW) \citep{robins1999association} and incorporate the state-of-the-art conditional generative models such as the guided diffusion model and conditional variational autoencoder.
We evaluate our framework through numerical experiments extensively on both synthetic and real-world data sets. 
\subsection{Related work}

Our work has connections to causal inference in time series, counterfactual density estimation, and generative models. To our best knowledge, our work is the first to intersect the three aforementioned areas. Below we review each of these areas independently.

\paragraph{Causal inference with time-varying treatments}
Causal inference has historically been related to longitudinal data. Classic approaches to analyzing time-varying treatment effects include the g-computation formula, structural nested models, and marginal structural models \citep{rubin1978bayesian, robins1986new, robins1994correcting,robins1994estimation, robins2000marginal, fitzmaurice2008longitudinal,li2021g}. These seminal works are typically based on parametric models with limited flexibility. Recent advancements in machine learning have significantly accelerated progress in this area using flexible statistical models \citep{schulam2017reliable,chen2023multi} and deep neural networks \citep{lim2018forecasting,bica2019estimating, berrevoets2021disentangled,li2021g, seedat2022continuous, melnychuk2022causal, frauen2023estimating, vanderschueren2023accounting} to capture the complex temporal dependency of the outcome on treatment and covariate history. These approaches, however, focus on predicting the mean counterfactual outcome instead of the distribution. The performance of these methods also heavily relies on the specific structures (\eg, long short term memory) without more flexible architectures.

\paragraph{Counterfactual distribution estimation}
Recently, several studies have emerged to estimate the entire counterfactual distribution rather than the means, including estimating quantiles of the cumulative distributional functions (CDFs) \citep{chernozhukov2013inference, wang2018quantile}, re-weighted kernel estimations \citep{10.2307/2171954}, and semiparametric methods \citep{Kennedy23}.  In particular, 
\cite{Kennedy23} highlights the extra information afforded by estimating the entire counterfactual distribution and using the distance between counterfactual densities as a measure of causal effects. \cite{melnychuk2023normalizing} uses normalizing flow to estimate the interventional density. However, these methods are designed to work under static settings with no time-varying treatments \citep{alaa2017bayesian}, and are explicit density estimation methods that may be difficult to scale to high-dimensional outcomes. \cite{li2021g} proposes a deep framework based on G-computation which can be used to simulate outcome trajectories on which one can estimate the counterfactual distribution. However, this framework approximates the distribution via empirical estimation of the sample variance, which may be unable to capture the complex variability of the (potentially high-dimensional) distributions. Our work, on the other hand, approximates the counterfactual distribution with a generative model without explicitly estimating its density. This will enable a wider range of application scenarios including continuous treatments and can accommodate more intricate data structures in the high-dimensional outcome settings.

\paragraph{Counterfactual generative model}
Generative models, including a variety of deep network architectures such as generative adversarial networks (GAN) and autoencoders, have been recently developed to perform counterfactual prediction \citep{goudet2017causal,louizos2017causal,yoon2018ganite,saini2019multiple, sauer2021counterfactual,van2021conditional, im2021causal, kuzmanovic2021deconfounding, balazadeh2022partial,fujii2022estimating,liu2022causalegm, reynaud2022d,zhang2022exploring}.
However, many of these approaches primarily focus on using representation learning to improve treatment effect estimation rather than obtaining counterfactual samples or approximating counterfactual distributions. For example, \cite{yoon2018ganite, saini2019multiple} adopt deep generative models to improve the estimation of individual treatment effects (ITEs) under static settings. Some of these approaches focus on exploring causal relationships between components of an image \citep{sauer2021counterfactual, van2021conditional, reynaud2022d}. Furthermore, there has been limited exploration of applying generative models to time series settings in the existing literature. 
A few attempts, including \cite{louizos2017causal, kuzmanovic2021deconfounding}, train autoencoders to estimate treatment effect using longitudinal data. 
Nevertheless, these methods are not intended for drawing counterfactual samples. In sum, to the best of our knowledge, our work is the first to use generative models to approximate counterfactual distribution from data with time-varying treatments, a novel setting not addressed by prior works.

\section{Methodology}

\subsection{Problem setup}
\begin{table}[t!]
 \caption {Summary of key notations}
\vspace{-.1in}
 \centering
 \begin{adjustbox}{max width=.7\textwidth}
 \begin{threeparttable}
 \begin{tabular}{ c   c   c } 
\toprule[1pt]\midrule[0.3pt]

\textbf{Variable} & \textbf{Domain}& \textbf{Description}\\
\hline
$t$ & $\{1,2,\cdots, T\}$ &  Time index \\
$d$ & $\mathbb{Z}^+$ &  Length of the history dependence \\
$z$ & $\mathbb{R}^{r}$ & Random noise vector  \\
$h$ & $\mathbb{Z}^+$ &  Covariate dimension \\
$m$ & $\mathbb{Z}^+$ &  Outcome dimension \\
$A_t$ & $\{0,1\}$ &  Treatment at time $t$  \\
$X_t$ & $\mathbb{R}^{h}$ &  Covariate at time $t$  \\
$Y_t$ & $\mathbb{R}^{m}$ &  Outcome at time $t$  \\
$\AB_t$ & $\{0,1\}^d$ &  Treatment history at time $t$  \\
$\XB_t$ & $\mathbb{R}^{d\times h}$ &  Covariate history at time $t$  \\
$ f(y, \ab, \xb)$ & $\mathbb{R}^+$ & Joint density\\
$ f_{\ab}(y)$ & $\mathbb{R}^+$ & Counterfactual density\\
$g_{\theta}(z,\ab)$& $\mathbb{R}^{m}$ & Counterfactual generator\\
$w_{\phi}(\ab,\xb)$& $\mathbb{R}^{+}$ & IPTW score\\

\midrule[0.3pt]\bottomrule[1pt]
\end{tabular}
\end{threeparttable}
 \end{adjustbox}
 \label{tab:notations}
\end{table}

In this study, we consider the treatment for each discrete time period (such as day or week) as a random variable $A_t \in \mathcal{A} = \{0,1\}$, where $t=1,\dots,T$ and $T$ is the total number of time points. Note that our framework also works with categorical and even continuous treatments. Let $X_t$ $\in \mathcal{X} \subset \mathbb{R}^{h}$ be the time-varying covariates, and $Y_t$ $\in \mathcal{Y} \subset \mathbb{R}^{m}$ the subject's outcome at time $t$. We use $\AB_t= \{A_{t-d+1}, \dots, A_{t}\}$ to denote the previous treatment history from time $t-d+1$ to $t$, where $d$ is the length of history dependence. Similarly, we use $\XB_t = \{X_{t-d+1},\dots,X_{t}\}$ to denote the covariate history. We use $y_t$, $a_t$, and $x_t$ to represent a realization of $Y_t$, $A_t$, and $X_t$, respectively, and use $\ab_t = (a_{t-d+1}, \dots, a_t)$ and $\xb_t = (x_{t-d+1}, \dots, x_t)$ to denote the history of treatment and covariate realizations. In the sections below, we will refer to $Y_t$, $\AB_t$, and $\XB_t$ as simply $Y$, $\AB$, and $\XB$, where $t$ will be clear from context. \rebut{Since the outcome is independent conditioning on its history, we can consider the samples across time and individuals as conditionally independent tuples ($y^i$, $\ab^i$, $\xb^i$), where $i$ denotes the sample index and the sample size is $N$.} The notations are summarized in Table~\ref{tab:notations}. 

\begin{figure}[t!]
\centering
\includegraphics[width=.6\textwidth]{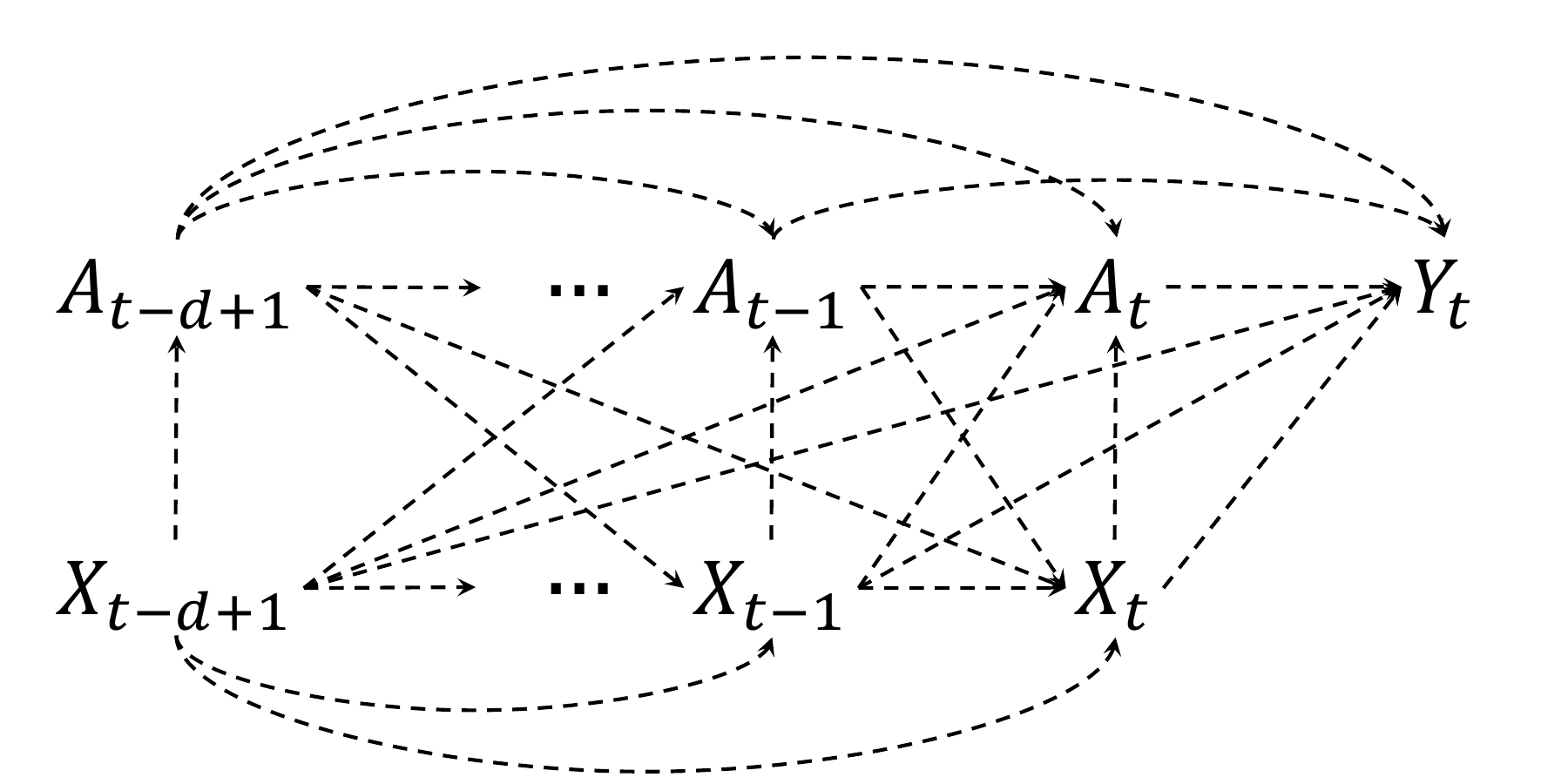}
\caption{The causal directed acyclic graph (DAG) of the time-varying treatment.}
\label{fig:graphical-model}
\end{figure}

The goal of our study is to obtain realistic samples of the counterfactual outcome \rebut{for all} given time-varying treatment $\ab$, without estimating its counterfactual density. 
Let $Y(\ab)$ denote the counterfactual outcome for a subject under a time-varying treatment $\ab$, and define $f_{\ab}$ as its counterfactual distribution. 
We note that $f_{\ab}$ is different from the marginal density of $Y$, as the treatment is fixed at $\AB = \ab$ in the conditioning. 
It is also not equal to the unadjusted conditional density $f(y | \ab)$. 
Instead, $f_{\ab}$ is the density of the counterfactual variable $Y(\ab)$, which represents the outcome that would have been observed if treatment were set to $\AB = \ab$. We assume the standard assumptions needed for identifying the treatment effects \citep{fitzmaurice2008longitudinal, lim2018forecasting, schulam2017reliable,imai2004causal}:
\begin{enumerate}
    \item \emph{Consistency}: If $\AB_t = \ab_t$ for a given subject, then the counterfactual outcome for treatment, $\ab_t$, is the same as the observed (factual) outcome: $Y(\ab_t) = Y$.
    \item \emph{Positivity}: If $\mathbb{P}\{\AB_{t-1} = \ab_{t-1}, \XB_t = \xb_t\} \neq 0$, then $\mathbb{P}\{\AB_t = \ab_t|\AB_{t-1} = \ab_{t-1}, \XB_t = \xb_t\}>0$ for all $\ab_t$.
    \item \emph{Sequential strong ignorability}: $Y(\ab_t) \indep \AB_t | \AB_{t-1}=\ab_{t-1}, \XB_t=\xb_t$, for all $a_t$ and $t$.  
\end{enumerate}
Assumption 2 means that, for each timestep, each treatment has a non-zero probability of being assigned. Assumption 3 (also called conditional exchangeability) means that there are no unmeasured confounders, that is, all of the covariates affecting both the treatment assignment and the outcomes are present in the the observational dataset. Note that while assumption 3 is standard across all methods for estimating treatment effects, it is not testable in
practice \citep{pearl2009causal, robins2000marginal}.
We also assume that $Y$, $\AB$, and $\XB$ follows the typically structural causal relationship as shown in Figure~\ref{fig:graphical-model}, which is a classical setting in longitudinal causal inference \citep{robins1986new,robins2000marginal}.

\subsection{Counterfactual generative framework for time-varying treatments}

This paper proposes a counterfactual generator, denoted as $g_\theta$, to simulate $Y(\ab)$ according to the \emph{proxy conditional distribution} $f_\theta(y|\ab)$ instead of directly modeling its expectation or specifying a parametric counterfactual distribution. 
Here we use $\theta \in \Theta$ to represent the model's parameters, and formally define the generator as a function:
\begin{equation}
    g_\theta(z, \ab): \R^r \times \mathcal{A}^d \rightarrow \mathcal{Y}.
    \label{eq:cond-generator}
\end{equation}
The generator takes as input a random noise vector ($z \in \R^r \sim \mathcal{N}(0, I)$) and the time-varying treatment $\ab$. Note that our selection of the generator $g$ is not constrained to a particular function. Instead, it can represent an arbitrary generative process, such as a diffusion model, provided it has the capability to produce a sample when supplied with noise and treatment history. In addition, the noise dimension, $r$, depends on the specific generator: for conditional variational autodencoders, it corresponds to the latent dimension, whereas for the guided diffusion models, it has the same dimensionality, $m$, as the outcome.

\begin{figure}[t!]
\centering
\includegraphics[width=0.7\textwidth]{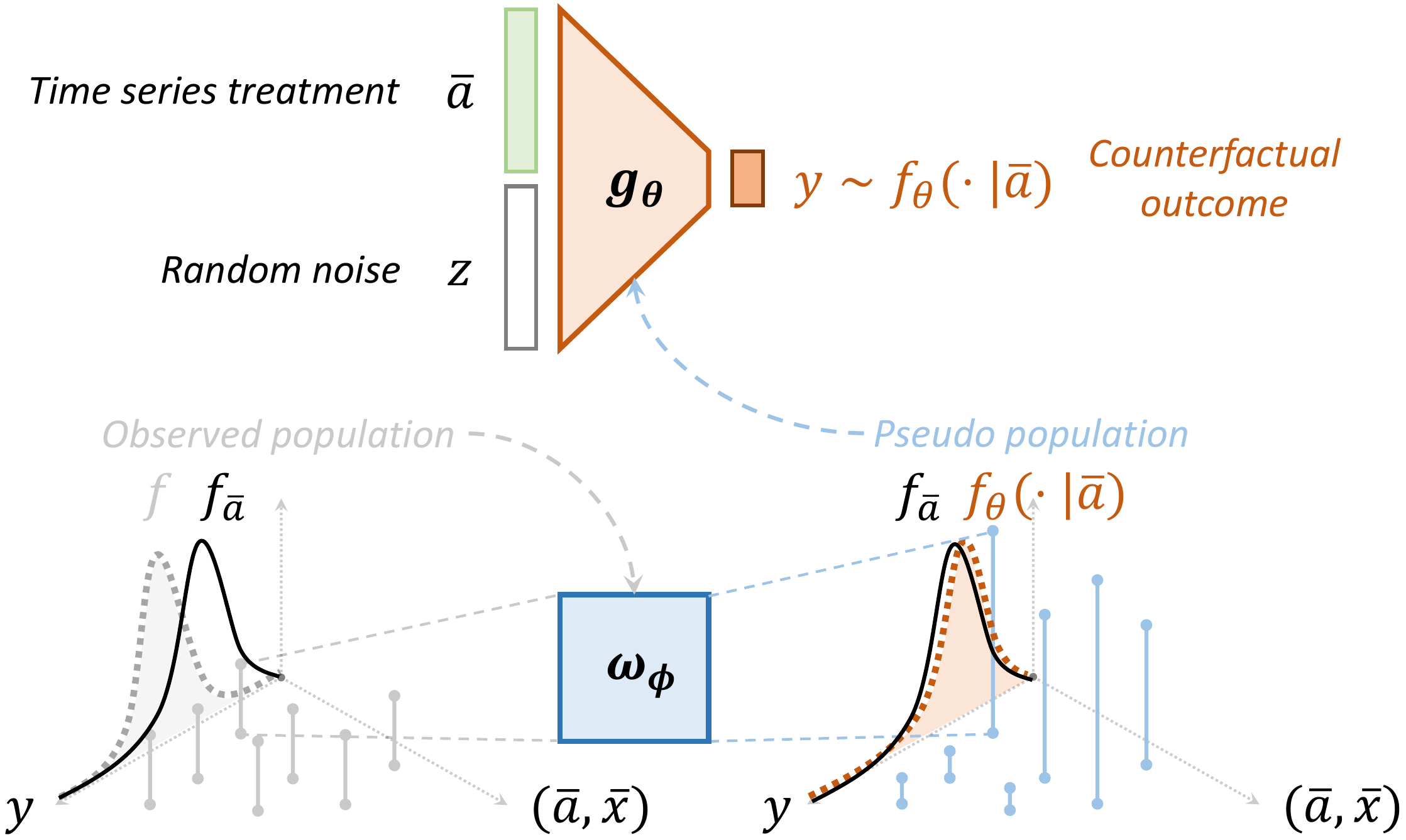}
\caption{The architecture of the proposed counterfactual generative models. The generator $g_\theta$ is designed to produce samples of the outcome variable $Y(\ab)$ with a given time-varying treatment $\ab$. The generated samples are expected to conform to the proxy conditional distribution $f_{\theta}$, which is an approximate of the underlying counterfactual distribution $f_{\ab}$. 
}
\label{fig:model-architecture}
\end{figure}

The output of the generator is a sample of possible counterfactual outcomes that follows the proxy conditional distribution represented by $\theta$, \ie,
\[
    y \sim f_{\theta}(\cdot|\ab),
\]
which can be viewed as an approximate of the underlying counterfactual distribution $f_{\ab}$.
Figure~\ref{fig:model-architecture} shows an overview of the proposed generative model architecture. 

\begin{figure}[!t]
\centering
\includegraphics[width=.6\textwidth]{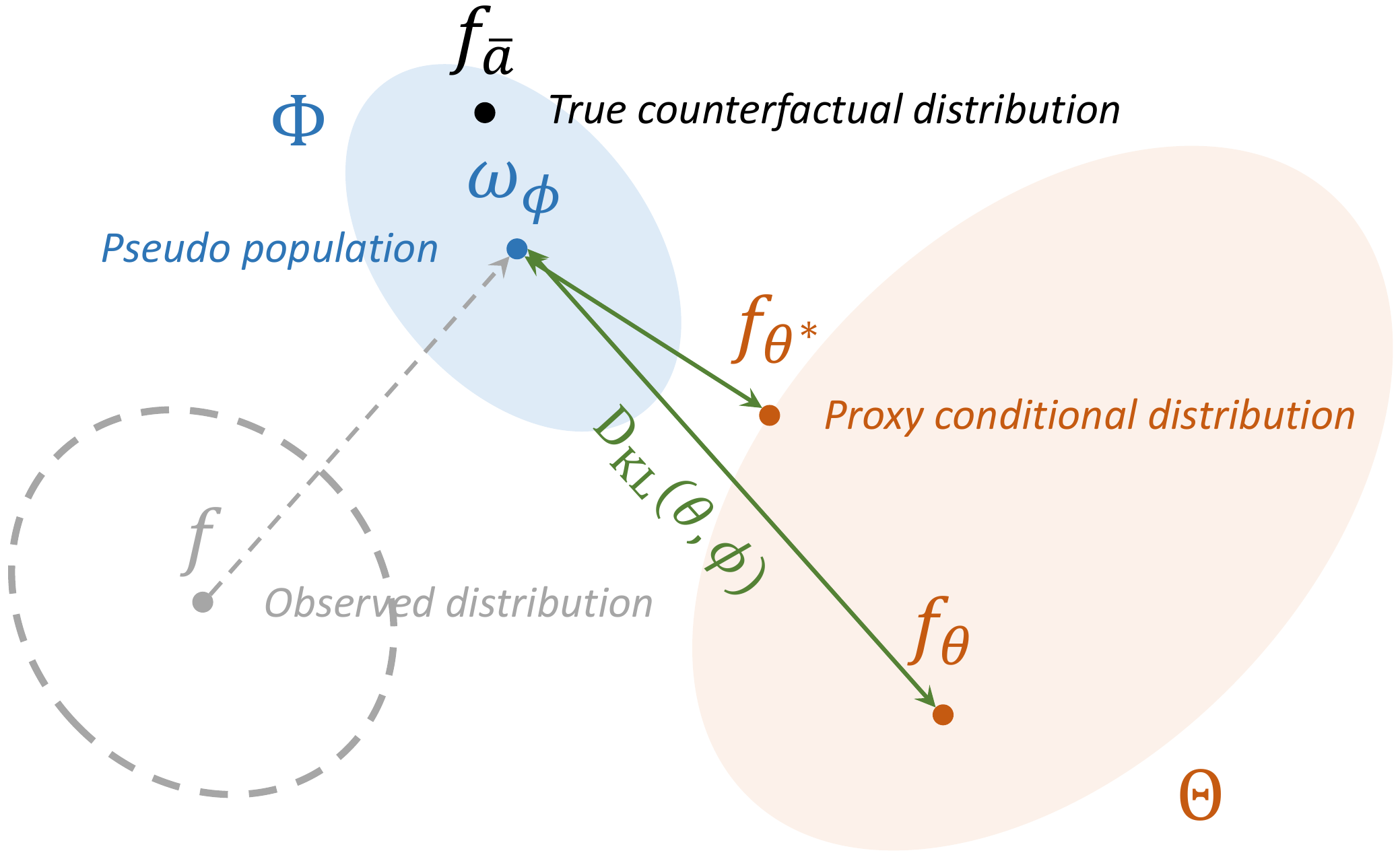}
\caption{An illustration of our learning objective. We aim to minimize the KL-divergence between the proxy distribution $f_\theta(\cdot|\ab)$ and the true counterfactual distribution $f_{\ab}$. 
}
\label{fig:objective}
\end{figure}

\paragraph{Marginal structural generative models}
The learning objective is to find the optimal generator that minimizes the distance between the proxy conditional distribution $f_\theta(\cdot|\ab)$ and the true counterfactual distribution $f_{\ab}$ \rebut{for any treatment sequence, $\ab$}, as illustrated in Figure~\ref{fig:objective}. For a general distributional distance, $D_f(\cdot,\cdot)$, the objective can be expressed as finding an optimal $\hat{\theta}$ that minimizes the difference between $f_\theta(\cdot|\ab)$ and $f_{\ab}$ over all treatments $\ab$ (\ie, over a uniform distribution of $\AB$):
\begin{equation}
    \hat{\theta} 
    = \argmin_{\theta \in \Theta}~\EE_{\AB } ~\left[D_f(f_{\ab}, f_\theta(\cdot|\ab)) \right].
    \label{eq:obj-general}
\end{equation}

If the distance metric is Kullback-Leibler (KL) divergence, this objective can be expressed equivalently by maximizing the log-likelihood \cite{murphy2012machine}. The proof can be found in Appendix~\ref{append:proof-obj}:
\begin{equation}
    \hat{\theta} 
    = \argmax_{\theta \in \Theta}~\EE_{\AB} ~\left[\EE_{y \sim f_{\ab}} ~\log f_\theta(\cdot|\ab)\right].
    \label{eq:log-likelihood}
\end{equation}

To obtain samples from the counterfactual distribution $f_{\ab}$, we follow the idea of marginal structural models (MSMs) introduced by \cite{neyman1923applications, rubin1978bayesian, robins1999association} and extended by \cite{fitzmaurice2008longitudinal} to account for time-varying treatments.
Specifically, we introduce Lemma~\ref{lemma:pseudo-prob}, which follows the g-formula proposed in \cite{robins1999association} and establishes a connection between the counterfactual distribution and the data distribution. 
The proof can be found in Appendix~\ref{append:proof-lemma-1}. 

\begin{lemma}
    Under unconfoundedness and positivity, we have
    \begin{equation}
        \cprob{\ab}{y}  =\int \frac{1}{\prod_{\tau=t-d+1}^{t} \prob{a_{\tau}|\ab_{\tau-1},\xb_{\tau}}} \prob{y,\ab,\xb} d\xb
    \end{equation}
    where $f(y,\ab,\xb)$ denotes the joint distribution and $\prob{a_{\tau}|\ab_{\tau-1},\xb_{\tau}}$ denotes the propensity score at $\tau$. 
    \label{lemma:pseudo-prob}
\end{lemma}


Now we present a proposition using Lemma~\ref{lemma:pseudo-prob}, allowing us to substitute the expectation in (\ref{eq:log-likelihood}), computed over a counterfactual distribution, with the sample average over a pseudo-population.
This pseudo-population is constructed by assigning weights to each data tuple based on their subject-specific IPTW. Figure~\ref{fig:objective} gives an illustration of the learning objective.
See the proof in Appendix~\ref{append:proof-prop-1}. 

\begin{prop}
    Let $\mathcal{D}$ denote the set of observed data tuples. 
    The generative learning objective can be approximated by:
    \begin{equation}
        \EE_{\AB} ~\left[\EE_{y \sim f_{\ab}} ~\log f_\theta(y|\ab)\right] \approx
        \frac{1}{N}\sum_{(y,\ab,\xb) \in \mathcal{D}}w_\phi(\ab,\xb) \log f_\theta(y|\ab),
        \label{eq:weighted-log-likelihood}
    \end{equation}
    where $N$ represents the sample size, and  $w_\phi(\ab,\xb)$ denotes the subject-specific IPTW, parameterized by $\phi \in \Phi$, which takes the form:
    \begin{equation}
        w_\phi(\ab,\xb) = \frac{1}{\prod_{\tau=t-d+1}^{t} f_\phi(a_{\tau}|\ab_{\tau-1},\xb_{\tau})}.
        \label{eq:iptw}
    \end{equation}
    \label{prop:pesudo-objective}
\end{prop}

\begin{rmk}
    \rebut{The generative learning objective in Proposition~\ref{prop:pesudo-objective} offers specific benefits when compared to plugin methods using Lemma~\ref{lemma:pseudo-prob} \citep{bickel2001inference,kim2018causal} and density estimators \citep{melnychuk2023normalizing}. \rebutkdd{We only used IPTW in the training of generative models, instead of combining an outcome model with IPTW like the widely used doubly robust method, due to the challenges of learning an outcome model for high-dimensional counterfactual distributions.} We include a detailed discussion in Appendix~\ref{append:connection-other}}. 
\end{rmk}

\rebutkdd{Our Proposition 1 intuitively posits that to effectively approximate the counterfactual distribution using a generative model, one can adjust the weighting of samples based on their IPTW during the model's training phase. This reweighting process debiases the observed outcome distribution to match its counterfactual distribution, thereby enhancing the model's ability to generate accurate counterfactual distributions.} Here we use another model, denoted by $\phi \in \Phi$, to represent the propensity score $f_\phi(a_{\tau}|\ab_{\tau}, \xb_{\tau})$, which defines the IPTW $w_\phi$ and can be learned separately using observed data. 
Note that the effectiveness of this method is dependent on a correct specification of the IPTW $\phi$, \ie, the black dot is inside of the blue area in Figure~\ref{fig:objective} \citep{fitzmaurice2008longitudinal, lim2018forecasting}. 
In \cite{lim2018forecasting}, they use an RNN-like structure to represent the conditional probability $f_\phi$ without making strong assumptions on the form of the conditional probability.
The choice of $g_\theta$ and $f_\phi$ are entirely general and both can be specified using deep architectures.
In this paper, we use fully-connected neural networks for both $g_\theta$ and $f_\phi$.


\begin{algorithm}[t!]
\begin{algorithmic}
    \STATE {\bfseries Input:} 
    Training set $\mathcal{D}$ data tuples: \rebut{$\mathcal{D} = \{(y^{i}, \ab^{i}, \xb^{i})\}_{i=1,\dots,N}$ where $N$ is the total number of training samples;}  the number of learning epoch $E$.
    \STATE {\bfseries Initialization:} model parameters $\theta$ and fitted $\widehat \phi$ using $\mathcal{D}$.
    \WHILE{$e<E$}
        \FOR{each sampled batch \rebut{$\mathcal{B}$} with size $n$}
        \STATE 1. Draw samples $\epsilon \sim \mathcal{N}(0, I)$ from noise distribution, whose dimensionality is either the outcome dimension (for the guided diffusion) or the latent dimension (for CVAE); $I$ is the identity matrix;
        \STATE 2. Compute the loss for $(y, \ab, \xb) \in$ \rebut{$\mathcal{B}$} given $\epsilon$ and $\theta$ either according to (\ref{eq:diffusion-lower-bound}) or (\ref{eq:variational-lower-bound});
        \STATE 3. Re-weight the loss for $(y, \ab, \xb) \in$ \rebut{$\mathcal{B}$} using $w_{\widehat \phi}(\ab, \xb)$ according to (\ref{eq:iptw});
        \STATE 4. Update $\theta$ by maximizing (\ref{eq:weighted-log-likelihood}) with gradient descent. 
        \ENDFOR
    \ENDWHILE
    \STATE {\bfseries return} $\theta$ \;
    
\end{algorithmic}
\caption{Learning algorithm for the conditional generator $\theta$}
\label{algo:learning}
\end{algorithm}

To maximize the objective as expressed in (\ref{eq:weighted-log-likelihood}) and learn the proposed generative model, one needs to compute the conditional log-likelihood, $\log f_\theta(\cdot|\ab)$ for any $\ab$, which usually has no closed-form. We can leverage various state-of-the-art generative learning algorithms that approximate the likelihood. To demonstrate the flexibility of our proposed marginal structural generative framework, we explore two state-of-the-art models: the guided diffusion models \citep{dhariwal2021diffusion,ho2022classifier} and the conditional variational autoencoder (CVAE) \citep{sohn2015learning}. 
We summarize our learning procedure in Algorithm~\ref{algo:learning}. 

\paragraph{Classifier-free guided diffusion model}

Diffusion models have been commonly used to generate realistic samples in domains such as images \citep{yang2023diffusion, ho2020denoising}. Here we adopt the classifier-free guidance framework \citep{ho2022classifier} by predicting the noise conditioning on the treatment:
\begin{equation}
  \log f_\theta(\cdot|\ab) \geq  -\mathbb{E}_{s \sim [1,S],y\sim f(y|\ab),\epsilon_s}||
  \epsilon_s - \epsilon_\theta(\sqrt{\bar{\lambda}_s}y+\sqrt{1-\bar{\lambda}_s}\epsilon_s,s,\ab)||^2,
    \label{eq:diffusion-lower-bound} 
\end{equation}
where $s$ denotes the denoising step, $S$ is the total number of steps, and $\epsilon_s$ is the Gaussian noise at the $s$-th step. Here $\bar{\lambda}_s=\prod_{s=1}^S(1-\gamma)$, where $\gamma$ is the noise variance at $s$ and $ \epsilon_\theta$ is the score function which is parameterized by a neural network to predict the noise at step $s$. Note that with the classifier-free guidance, $\epsilon_\theta$, is a linear combination of conditional and unconditional score functions (see Appendix \ref{app:conditional-ddpm}).

\paragraph{Conditional variational autoencoder}
For the CVAE, we approximate the logarithm of the proxy conditional probability using its evidence lower bound represented with an encoder-decoder form:
\begin{equation}
   \log f_\theta(\cdot|\ab) \geq -D_\text{KL}\left(q(z|y, \ab)||p_{\theta}(z|\ab)\right) + \mathbb{E}_{q_\theta(z|y, \ab)}\left[\log p_{\theta}(y|z, \ab) \right],
    \label{eq:variational-lower-bound} 
\end{equation}
where $p_{\theta}(y|z, \ab)$ is our conditional generator, and $q_\theta(z|y, \ab)$ and $p_{\theta}(z|\ab)$ are both parameterized by neural networks as encoder and prior. The complete derivation of (\ref{eq:variational-lower-bound}) and implementation details can be found in Appendix~\ref{append:variational-learning}.

\section{Experiments}
We evaluate our method using numerical examples and demonstrate the superior performance compared to the state-of-the-art methods. These are (1) Kernel Density Estimation (\texttt{KDE}) \citep{rosenblatt1956remarks}, (2) Marginal structural model with a fully-connected neural network (\texttt{MSM+NN}) \citep{robins1999estimation, lim2018forecasting}, (3) Conditional Variational Autoencoder (\texttt{CVAE}) \citep{sohn2015learning}, (4) Semi-parametric Plug-in method based on pseudo-population (\texttt{Plugin+KDE}) \citep{kim2018causal}, and (5) G-Net (\texttt{G-Net}) \citep{li2021g}. In the following, we refer to our proposed conditional generator as marginal structural conditional variational autoencoder (\texttt{MSCVAE}) and marginal structural diffusion (\texttt{MSDiffusion}) to show the flexibility of our generative framework. 
Given that the diffusion model with classifier-free guidance primarily targets high-dimensional data, such as images \citep{ho2022classifier,rombach2022high,yang2023diffusion}, we focus the evaluation of \texttt{MSDiffusion} on semi-synthetic datasets with high outcome dimensionality (Section 3.2). Additionally, we introduce an unweighted \texttt{Diffusion} model \citep{ho2020denoising} as our baseline for such comparisons. We also compare to the Counterfactual Recurrent Network (\texttt{CRN})\citep{bica2019estimating} on fully synthetic data.
Here, the \texttt{G-Net} is based on G-computation. The \texttt{Plugin+KDE} is tailored for counterfactual density estimation. The \texttt{CVAE} and \texttt{Diffusion} act as a reference model, highlighting the significance of IPTW reweighting in our approach. See Appendix~\ref{append:baselines} for a detailed review of these baseline methods. 

\paragraph{Experiment set-up}
To learn the model parameter $\theta$, we use stochastic gradient descent to maximize the weighted log-likelihood (\ref{eq:weighted-log-likelihood}). 
We adopt an Adam optimizer \citep{kingma2014adam} with a batch size of $256$, a learning rate of $10^{-3}$ for \texttt{MSCVAE} and a learning rate of $10^{-4}$ for \texttt{MSDiffusion}. 
To ensure learning stability, we follow a commonly-used practice \citep{xiao2010accuracy,lim2018forecasting} that involves truncating the subject-specific IPTW weights at the $0.01$-th and $99.99$-th percentiles and normalizing them by their mean. 
All experiments are performed with 16GB RAM and a 2.6 GHz 6-Core Intel Core i7 CPU. More details of the experiment set-up can be found in Appendix~\ref{append:set-up}.

\begin{figure*}[!t]
    \centering
    \includegraphics[width=\textwidth]{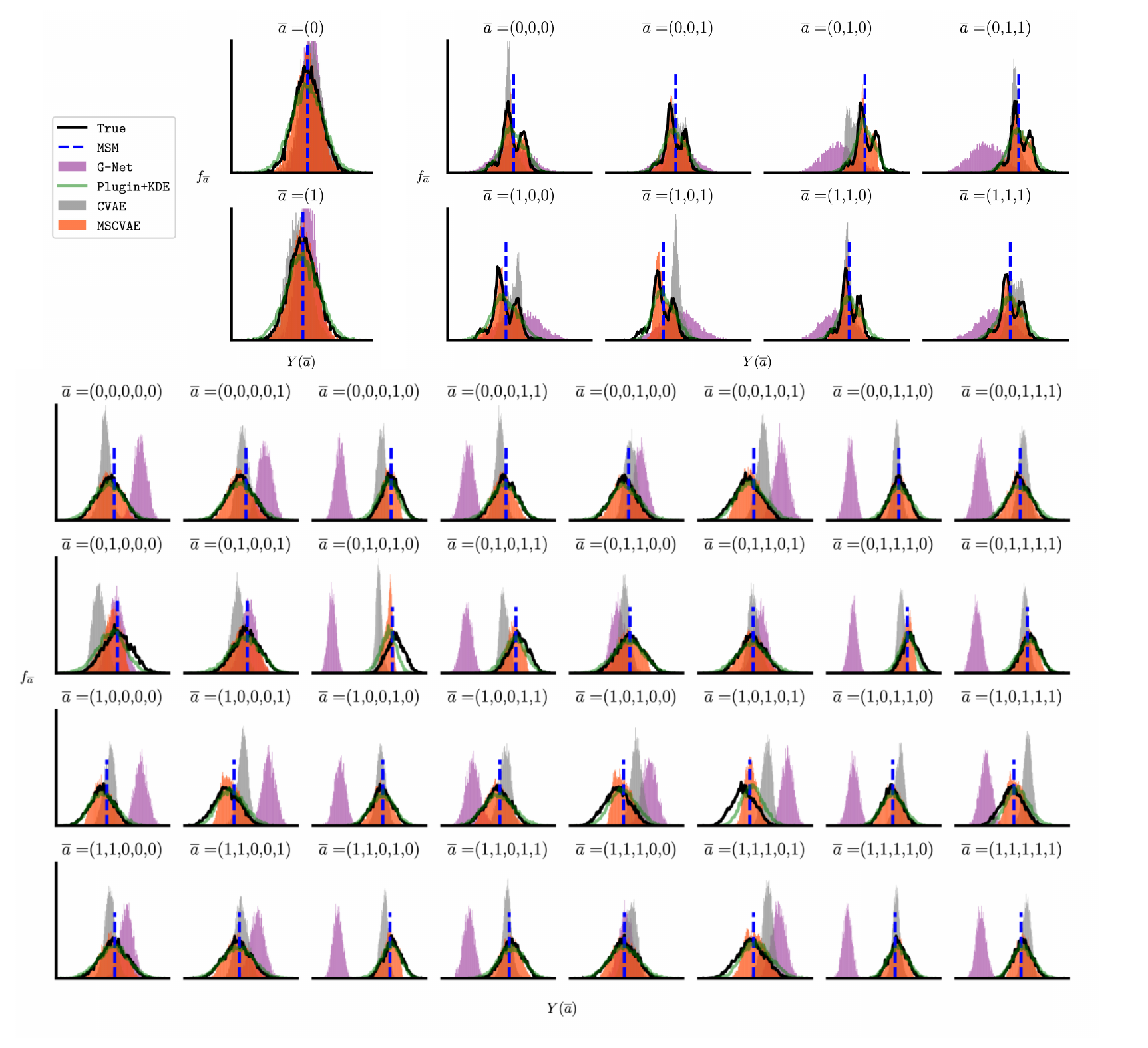}
    \vspace{-.2in}
\caption{The estimated and true counterfactual distributions across various lengths of history dependence ($d=1,3,5$) on the fully synthetic datasets ($m=1$). Each sub-panel provides a comparison for a specific treatment combination $\ab$.}
\label{fig:synthetic-exps}
\end{figure*}

\subsection{Fully synthetic data}

We first assess the effectiveness of the \texttt{MSCVAE} using fully synthetic experiments. The effectiveness of   \texttt{MSDiffusion} is evaluated on several semi-synthetic datasets in the next section  for the aforementioned reason.
Following the classical experimental setting described in \cite{robins1999estimation}, we simulate three synthetic datasets with different lengths of history dependence ($d=1,3,5$) using linear models. 
Each dataset comprises $10,000$ trajectories, representing recorded observations of individual subjects. These trajectories consist of $100$ data tuples, encompassing treatment, covariate, and outcome values at specific time points. See Appendix~\ref{append:synthetic} for a detailed description of the synthetic data generation. The causal dependence between these variables is visualized in Figure~\ref{fig:graphical-model}. In Figure~\ref{fig:synthetic-exps}, the \texttt{MSCVAE} (orange shading) outperforms the baseline methods in accurately capturing the shape of the true counterfactual distributions (represented by the black line) across all scenarios.
It is worth mentioning that the learned distribution produced by \texttt{CVAE} deviates significantly from the desired target, emphasizing the significance of the weighting term in Proposition \ref{prop:pesudo-objective} in accurately approximating the counterfactual distribution.

\begin{table*}[t!]
 \caption {Quantitative performance on fully-synthetic data}
 \centering
 \begin{adjustbox}{max width=0.9\textwidth}
 \begin{threeparttable}
 \begin{tabular}{ c   c   c  c c  c c  c  c} 
\toprule[1pt]\midrule[0.3pt]
\textbf{} &  \multicolumn{2}{c}{$d=1$} &  \multicolumn{2}{c}{$d=3$}& \multicolumn{2}{c}{$d=5$}  \\
\textbf{Methods} & {Mean $\downarrow$}& {Wasserstein $\downarrow$} &  {Mean $\downarrow$}& {Wasserstein $\downarrow$} & {Mean $\downarrow$}& {Wasserstein $\downarrow$}\\
\hline
\texttt{MSM+NN} & $\textbf{0.001}$ ($\textbf{0.002}$) &  $0.601$ ($0.603$) & $0.070$ ($\underline{0.159}$) &  $0.689$ ($0.718$)  & $0.198$ ($\textbf{0.563}$) &  $0.600$ ($0.737$)  \\

\texttt{KDE} & $0.246$ ($0.267$) &  $0.244$ ($0.268$) & $0.520$  ($1.080$)&  $0.538$ ($1.080$) & $0.538$ ($1.419$)  &  $0.539$ ($1.419$)  \\

\texttt{Plugin+KDE} & $0.010$ ($0.014$) &  $\textbf{0.034}$ ($\textbf{0.036}$) & $\textbf{0.045}$ ($0.168$) &  $\underline{0.132}$ ($\textbf{0.168}$)  & $\textbf{0.147}$ ($0.598$)  & $\underline{0.182}$ ($\textbf{0.598}$)  \\

\texttt{CRN} & $0.228$ ($0.280$)  & $0.289$ ($0.331$) & $0.913$ ($1.753$) &  $1.014$ ($1.757$)  & $1.713$ ($4.080$)  & $1.775$ ($4.080$)  \\

\texttt{G-Net} & $0.211$ ($0.258$)  & $0.572$ ($0.582$) & $1.167$ ($2.173$) &  $1.284$ ($2.173$)  & $2.314$ ($5.263$)  & $2.354$ ($5.263$)  \\

\texttt{CVAE} & $0.250$ ($0.287$) &  $0.253$ ($0.288$) & $0.517$ ($1.061$) &  $0.553$ ($1.061$) & $0.539$ ($1.430$) &  $0.613$ ($1.430$)\\

\textcolor{orange}{\texttt{MSCVAE}} & $\underline{0.006}$ ($\underline{0.006}$) &  $\underline{0.055}$ ($\underline{0.056}$) & $\underline{0.046}$ ($\textbf{0.150}$) &  $\textbf{0.105}$  ($\underline{0.216}$) &$\underline{0.150}$ ($\underline{0.633}$) &  $\textbf{0.173}$ ($\underline{0.633}$)    \\

\textcolor{orange}{\texttt{MSDiffusion}} & $0.029$ ($0.052$) &  $0.056$ ($0.065$) & $0.086$ ($0.234$) &  $0.135$  ($0.234$) &$0.207$ ($0.845$) &  $0.259$ ($0.845$)    \\

\midrule[0.3pt]\bottomrule[1pt]
\end{tabular}
\begin{tablenotes}[para,flushleft]
      * Numbers represent the \textbf{average} metric across all treatment combinations and those in the parentheses represent the \textbf{worst} across treatment combinations. Bold and underlined numbers represent the best and second best results.
\end{tablenotes}
\end{threeparttable}
 \end{adjustbox}
 \label{tab:comparision}
\end{table*}

\begin{table}[t!]
 \caption {Quantitative performance on fully-synthetic data with imbalanced treatment and conditional outcome}
 \centering
 \begin{adjustbox}{max width=.8\textwidth}
 \begin{threeparttable}
 \begin{tabular}{ c   c   c c  c } 
\toprule[1pt]\midrule[0.3pt]
\textbf{} & \multicolumn{2}{c}{\textbf{Imbalanced}} & \multicolumn{2}{c}{\textbf{Conditional}}\\
\textbf{} & \multicolumn{4}{c}{$d=5$}   \\
\textbf{Methods}  & Mean $\downarrow$ &Wasserstein  $\downarrow$  & Mean $\downarrow$ &Wasserstein  $\downarrow$ \\
\hline
\texttt{MSM+NN} & $0.173$ ($\textbf{0.448}$)&  $0.502$ ($\textbf{0.613}$) & $\textbf{0.164}$ ($\textbf{0.441}$) & $0.368$ ($\textbf{0.517}$) \\

\texttt{KDE} & $0.518$ ($1.504$) & $0.520$ ($1.504$) & $0.562$ ($1.590$) & $0.564$ ($1.590$)\\

\texttt{Plugin+KDE} & $\textbf{0.157}$ ($0.863$)  & $\underline{0.211}$ ($0.863$) & $0.170$ ($0.823$) & $\underline{0.196}$ ($0.823$) \\

\texttt{G-Net} & $2.070$ ($4.794$) &$2.072$ ($4.794$) & $0.815$ ($2.238$) & $0.843$ ($2.238$)\\

\texttt{CVAE} & $0.521$ ($1.540$) &  $0.565$ ($1.540$) & $0.534$ ($1.478$) & $0.585$ ($1.478$)\\

\textcolor{orange}{\texttt{MSCVAE} }& $\underline{0.162}$ ($\underline{0.832}$) &  $\textbf{0.187}$ ($\underline{0.832}$)  & $\underline{0.169}$ ($\underline{0.767}$) & $\textbf{0.186}$ ($\underline{0.767}$) \\

\midrule[0.3pt]\bottomrule[1pt]
\end{tabular}
\begin{tablenotes}[para,flushleft]
Results for $d=1$ and $3$ as well as the visualizations are included in Appendix \ref{append:additional}.
\end{tablenotes}
\end{threeparttable}
 \end{adjustbox}
 \label{tab:comparision-additional}
\end{table}

\begin{table}[t!]
 \caption {Quantitative performance on semi-synthetic data}
 \centering
 \begin{adjustbox}{max width=0.5\textwidth}
 \begin{threeparttable}
 \begin{tabular}{ c   c   c } 
\toprule[1pt]\midrule[0.3pt]
\textbf{} & \textbf{COVID-19 }&\textbf{TV-MNIST}\\
\textbf{} & $m=67$ & $m=784$  \\
\textbf{Methods}  & FID* $\downarrow$ &FID*  $\downarrow$ \\
\hline
\texttt{MSM+NN} & $1.520$ ($2.434$)&  $1.236$ ($3.956$) \\

\texttt{KDE} & $1.689$ ($3.067$) & $1.509$ ($2.557$)\\

\texttt{Plugin+KDE} & $1.474$ ($1.584$)  & $1.370$ ($1.799$) \\

\texttt{G-Net} & $1.591$ ($2.804$) &$1.751$ ($6.096$)\\

\texttt{CVAE} & $0.916$ ($4.092$) &  $2.149$ ($5.484$)\\

\texttt{Diffusion} & 0.703 (2.971)  & 1.138 (2.891)\\

\textcolor{orange}{\texttt{MSCVAE} }& $\textbf{0.462}$ ($\textbf{0.838}$) &  $\textbf{0.270}$ ($\textbf{1.004}$)  \\

\textcolor{orange}{\texttt{MSDiffusion}}  & \underline{0.648} (\underline{0.918}) & \underline{0.734} (\underline{1.005})\\

\midrule[0.3pt]\bottomrule[1pt]
\end{tabular}
\end{threeparttable}
 \end{adjustbox}
 \label{tab:comparision-semisynthetic}
\end{table}

Table~\ref{tab:comparision} summarizes the quantitative comparisons across the baselines. For the fully synthetic datasets, we adopt two metrics: mean distance and $1$-Wasserstein distance \citep{frogner2015learning,panaretos2019statistical}, as commonly-used metrics to measure the discrepancies between the approximated and counterfactual distributions (see Appendix~\ref{append:metrics} for more details). The \texttt{MSCVAE} not only consistently achieves the smallest Wasserstein distance in the majority of the experimental settings, but also demonstrates highly competitive accuracy on mean estimation, which is consistent with the result in Figure~\ref{fig:synthetic-exps}. Note that even though our goal is not to explicitly estimate the counterfactual distribution, the results clearly demonstrate that our generative model can still accurately approximate the underlying counterfactual distribution, even compared to the unbiased density-based method such as \texttt{Plugin+KDE}.

To further compare the performance of the algorithms under extended application scenarios, we looked at two cases: when the dataset has imbalanced proportions of different treatment combinations and when there is a static baseline covariate for conditional counterfactual outcome generation. The \texttt{MSCVAE} consistently outperformed other baselines, as shown in Appendix \ref{append:additional}.

\subsection{Semi-synthetic data} 
\label{semi-synthetic}


To demonstrate the ability of our generative framework to generate credible high-dimensional counterfactual samples, we test both \texttt{MSCVAE} and \texttt{MSDiffusion} on two semi-synthetic datasets. 
The benefit of these datasets is that both factual and counterfactual outcomes are available. Therefore, we can obtain a sample from the ground-truth counterfactual distribution, which we can then use for benchmarking. 
We evaluate the performance by measuring the quality of generated samples and the true samples from the dataset.


\paragraph{Time-varying MNIST}
We create TV-MNIST, a semi-synthetic dataset using MNIST images \citep{deng2012mnist,jesson2021quantifying} as the outcome variable ($m=784$). In this dataset, images are randomly selected, driven by the result of a latent process defined by a linear autoregressive model, which takes a $1$-dimensional covariate and treatment variable as inputs and outputs a digit (between $0$ and $9$). 
Here we set the length of history dependence, $d$, to $3$. 
This setup allows us to evaluate the performance of the algorithms by visually contrasting the quality and distribution of generated samples against counterfactual ones.
The full description of the dataset can be found in Appendix ~\ref{append:tv-mnist}.

\begin{figure}[t!]
\centering
\includegraphics[width=\textwidth]{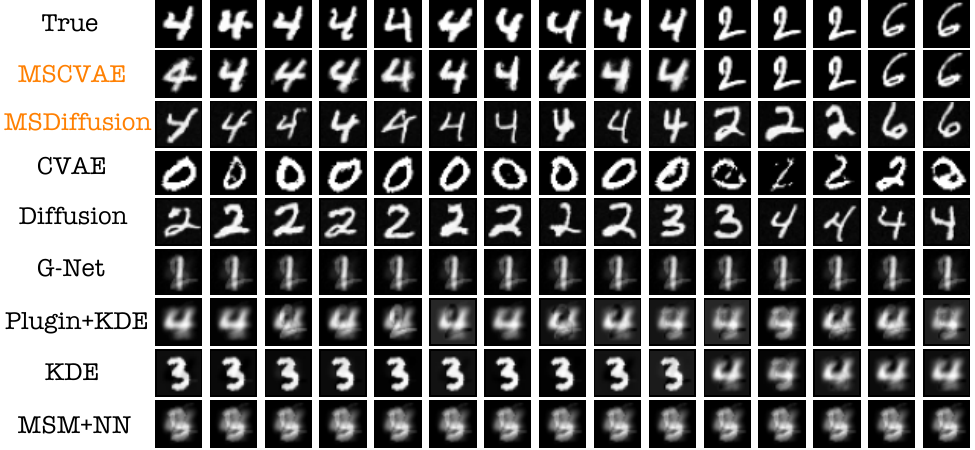}
\caption{Results on the semi-sythetic TV-MNIST datasets ($m=784$). We show representative samples generated from different methods under the treatment combinations $\ab=(1,1,1)$.}
\label{fig:tvmnist}
\end{figure}

\begin{figure*}[t!]
\centering
\includegraphics[width=\textwidth]{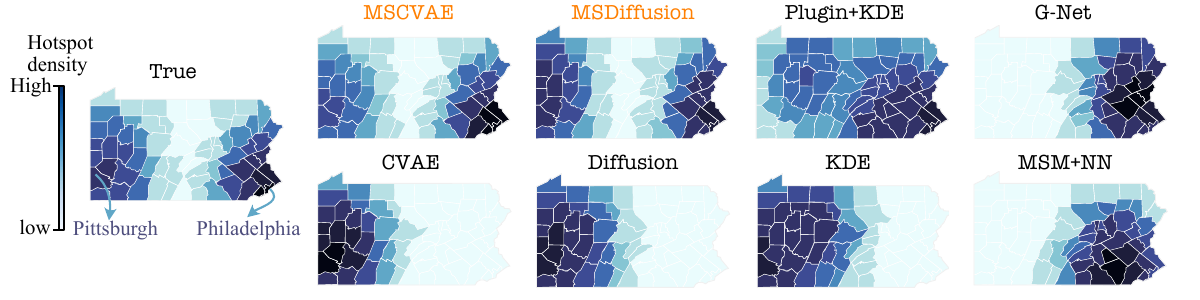}
\vspace{-.2in}
\caption{Results on the semi-synthetic Pennsylvania COVID-19 mask datasets ($m=67$) under the treatment combination $\ab=(1,1,1)$. We visualize the distribution of ``hotspots'' from the generated and true counterfactual distribution. For each model, we generate $500$ counterfactual samples. Each sample is a $67$-dimensional vector representing the inferred new cases per $100$K for the counties in Pennsylvania. We define the hotspot of each sample as the coordinate of the county with the highest number of new cases per $100$K, and visualize the density of the $500$ hotspots using kernel density estimation.}
\label{fig:semi-covid}
\end{figure*}

\paragraph{Pennsylvania COVID-19 mask mandate}
We create another semi-synthetic dataset to investigate the effectiveness of mask mandates in Pennsylvania during the COVID-19 pandemic.
We collected data from multiple sources, including the Centers for Disease Control and Prevention (CDC), the US Census Bureau, and a Facebook survey \citep{zhu2021high, centers2021us, zhu2022early, google, uscensus, delphi}. 
The dataset encompasses variables aggregated on a weekly basis spanning $106$ weeks from $2020$ and $2022$. There are four state-level covariates (per $100$K people): the number of deaths, the average retail and recreation mobility, the surveyed COVID-19 symptoms, and the number of administered COVID-19 vaccine doses. We set the state-level mask mandate policy (with values of $0$ indicating no mandate and $1$ indicating a mandate) as the treatment variable, and the county-level number of new COVID-19 cases (per $100$K) as the outcome variable ($m=67$). 
We simulate $2,000$ trajectories of the (covariate, treatment) tuples of $300$ time points (each point corresponding to a week) according to the real data. 
The outcome model is structured to exhibit a peak, defined as the "hotspot", in one of the state's two major cities: Pittsburgh or Philadelphia. The likelihood of these hotspots is contingent on the covariates. Consequently, the counterfactual and observed distributions manifest as bimodal, with varying probabilities for the hotspot locations. To ensure a pertinent analysis window, we've fixed the history dependence length, $d$, at $3$, aligning with the typical duration within which most COVID-19 symptoms recede \citep{maltezou2021post}.
The full description of the dataset can be found in Appendix ~\ref{append:real-data}. 

Given the high-dimensional outcomes of both semi-synthetic datasets, straightforward comparisons using means or the Wasserstein distance of the distributions tend to be less insightful. 
As a result, we use FID* (Fréchet inception distance *), an adaptation of the commonly-used FID \citep{heusel2017gans} to evaluate the quality of the counterfactual samples. For the TV-MNIST dataset, we utilize a pre-trained MNIST classifier, and for the COVID-19 dataset, a geographical projector, to map the samples into a feature space. 
Subsequently, we calculate the Wasserstein distance between the projected samples and counterfactual samples. 
The details can be found in Appendix ~\ref{append:metrics}. 


 From Figure~\ref{fig:tvmnist}, \ref{fig:semi-covid}, and Table~\ref{tab:comparision-semisynthetic}, we observe:
\begin{enumerate}
    \item Both the \texttt{MSCVAE} and \texttt{MSDiffusion}  outperform other baselines in generating samples that closely resemble the ground truth with overwhelmingly better FID* scores.

    \item Samples produced by the \texttt{Plugin+KDE} appear blurred in Figure~\ref{fig:tvmnist} and exhibit noise in Figure~\ref{fig:semi-covid}. This can be attributed to the inherent complexities of high-dimensional density estimation \citep{scott1983probability}, underscoring the value of employing a generative model to craft high-dimensional samples without resorting to precise density estimation. 

    \item  The superior results of \texttt{MSCVAE} (vs. \texttt{CVAE}), \texttt{MSDiffusion} (vs. \texttt{Diffusion}), and \texttt{Plugin+KDE} (vs. \texttt{KDE}) emphasize the pivotal role of IPTW correction during modeling.

    \item Deterministic approaches like \texttt{MSM+NN} fall short in capturing key features of the counterfactual distribution.

\end{enumerate}

In sum, the semi-synthetic experiments highlights the distinct 
benefits of our generative framework, 
particularly in generating high-quality counterfactual samples under time-varying treatments in a high-dimensional causal context.






\subsection{Real data}

\begin{figure}[t!]
    \centering
    \includegraphics[width=0.6\textwidth]{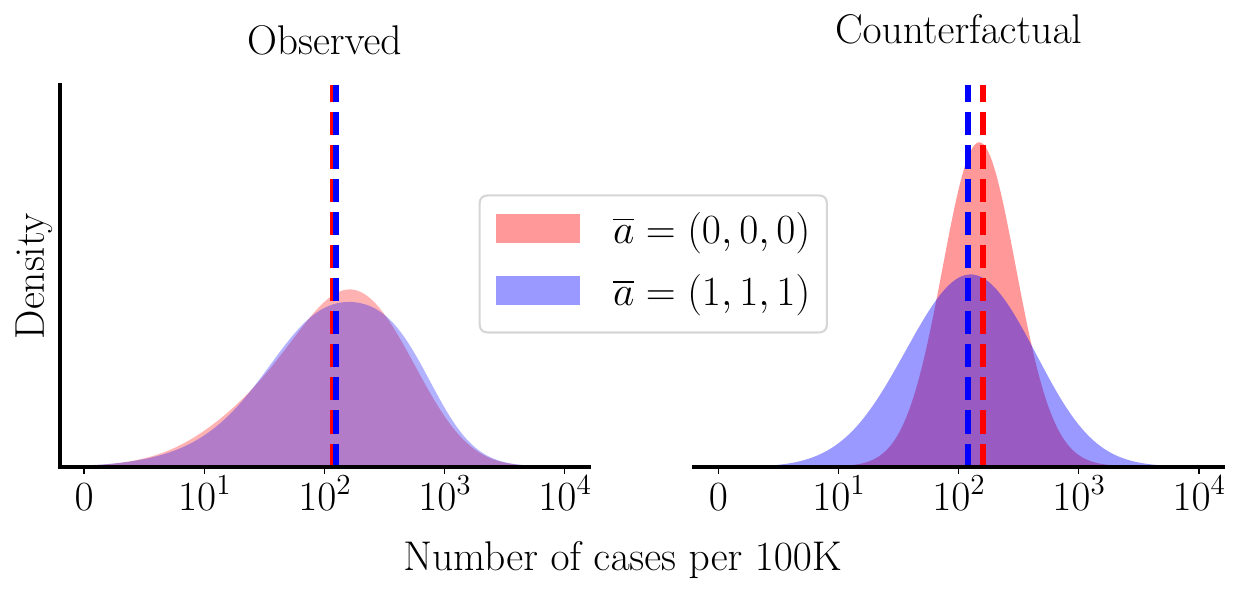}
    \caption{Observed distribution and estimated counterfactual distribution of the number of real COVID-19 cases per $100$K under two mask policies. The vertical dashed lines represent the mean of the corresponding distributions.}
    \label{fig:real-exps}
\end{figure}

We perform a case study using a real COVID-19 mask mandate dataset across the U.S. from $2020$ to $2021$ spanning $49$ weeks. We analyze the same set of variables as the semi-synthetic COVID-19 dataset except that: (1) We exclude the vaccine dosage from the covariates due to missing data in some states. (2) All variables in this dataset, including the treatments, the covariates, and the outcomes, are real data without simulation. Due to the limitation on the sample size for state-level observations, we only look at the county-level data, covering $3,219$ U.S. counties. This leads to $m=1$ and we utilize \texttt{MSCVAE} as our counterfactual generator. The details can be found in Appendix~\ref{append:real-data}.

Figure~\ref{fig:real-exps} illustrates a comparative analysis of the distribution of the observed and generated outcome samples under two different scenarios: one without a mask mandate ($\ab = (0, 0, 0)$) and the other with a full mask mandate ($\ab = (1, 1, 1)$).
In the left panel, we observe that the distributions under both policies appear remarkably similar, suggesting that the mask mandate has a limited impact on controlling the spread of the virus. This unexpected outcome challenges commonly held assumptions.
In the right panel, we present counterfactual distributions estimated using our method, revealing a noticeable disparity between the mask mandate and no mask mandate scenarios. 
The mean of the distribution for the mask mandate is significantly lower than that of the no-mask mandate. These findings indicate that implementing a mask mandate consistently for three consecutive weeks can effectively reduce the number of future new cases. It aligns with the understanding supported by health experts' suggestions and various studies \citep{van2020trends, adjodah2021association, guy2021association, nguyen2021mask, wang2021effective} regarding the effectiveness of wearing masks. Finally, it is important to note that the implementation of full mask mandates exhibits a significantly higher variance compared to the absence of a mask mandate. This implies that the impact of a mask mandate varies across different data points, specifically counties in our study. This insight highlights the need for policymakers to carefully assess the unique characteristics of their respective regions when considering the implementation of mask mandate policies. It is crucial for policymakers to understand that the effectiveness of a mask mandate may yield negative outcomes in certain extreme cases. Therefore, when proposing and implementing such policies, a thorough examination of the specific circumstances is highly recommended to avoid any unintended consequences.

\section{Conclusion}
We have introduced a powerful conditional generative framework tailored to generate samples that mirror counterfactual distributions in scenarios where treatments vary over time. 
Our model approximates the true counterfactual distribution by minimizing the KL-divergence between the true distribution and a proxy conditional distribution, approximated by generated samples. 
We have showcased our framework's superior performance against state-of-the-art methods in both fully-synthetic and real experiments.

Our proposed framework has great potential in generating intricate high-dimensional counterfactual outcomes and can be enhanced by adopting cutting-edge generative models and their learning algorithms, such as diffusion models. Additionally, our generative approach can be easily adapted to scenarios with continuous treatments, where the conditional generator enables extrapolation between unseen treatments under continuity assumptions.   

We also recognize potential caveats stemming from breaches in statistical assumptions. 
In real-world scenarios, the conditional exchangeability condition might be compromised due to unobserved confounders. 
Similarly, the positivity assumption could be at risk, attributed to the escalating number of treatment combinations as $d$ increases. 
Hence, meticulous assessment of these assumptions is imperative for a thorough and accurate statistical interpretation when employing our framework.

\section{Acknowledgement}
We would like to extend our sincere thanks to Liyan Xie for her valuable insights during our discussions.

\bibliography{refs.bib}


\begin{thebibliography}{89}


\ifx \showCODEN    \undefined \def \showCODEN     #1{\unskip}     \fi
\ifx \showDOI      \undefined \def \showDOI       #1{#1}\fi
\ifx \showISBNx    \undefined \def \showISBNx     #1{\unskip}     \fi
\ifx \showISBNxiii \undefined \def \showISBNxiii  #1{\unskip}     \fi
\ifx \showISSN     \undefined \def \showISSN      #1{\unskip}     \fi
\ifx \showLCCN     \undefined \def \showLCCN      #1{\unskip}     \fi
\ifx \shownote     \undefined \def \shownote      #1{#1}          \fi
\ifx \showarticletitle \undefined \def \showarticletitle #1{#1}   \fi
\ifx \showURL      \undefined \def \showURL       {\relax}        \fi
\providecommand\bibfield[2]{#2}
\providecommand\bibinfo[2]{#2}
\providecommand\natexlab[1]{#1}
\providecommand\showeprint[2][]{arXiv:#2}

\bibitem[Adjodah et~al\mbox{.}(2021)]%
        {adjodah2021association}
\bibfield{author}{\bibinfo{person}{Dhaval Adjodah}, \bibinfo{person}{Karthik
  Dinakar}, \bibinfo{person}{Matteo Chinazzi}, \bibinfo{person}{Samuel~P
  Fraiberger}, \bibinfo{person}{Alex Pentland}, \bibinfo{person}{Samantha
  Bates}, \bibinfo{person}{Kyle Staller}, \bibinfo{person}{Alessandro
  Vespignani}, {and} \bibinfo{person}{Deepak~L Bhatt}.}
  \bibinfo{year}{2021}\natexlab{}.
\newblock \showarticletitle{Association between COVID-19 outcomes and mask
  mandates, adherence, and attitudes}.
\newblock \bibinfo{journal}{\emph{PLoS One}} \bibinfo{volume}{16},
  \bibinfo{number}{6} (\bibinfo{year}{2021}), \bibinfo{pages}{e0252315}.
\newblock


\bibitem[Alaa and Van Der~Schaar(2017)]%
        {alaa2017bayesian}
\bibfield{author}{\bibinfo{person}{Ahmed~M Alaa} {and} \bibinfo{person}{Mihaela
  Van Der~Schaar}.} \bibinfo{year}{2017}\natexlab{}.
\newblock \showarticletitle{Bayesian inference of individualized treatment
  effects using multi-task gaussian processes}.
\newblock \bibinfo{journal}{\emph{Advances in neural information processing
  systems}}  \bibinfo{volume}{30} (\bibinfo{year}{2017}).
\newblock


\bibitem[Balazadeh~Meresht et~al\mbox{.}(2022)]%
        {balazadeh2022partial}
\bibfield{author}{\bibinfo{person}{Vahid Balazadeh~Meresht},
  \bibinfo{person}{Vasilis Syrgkanis}, {and} \bibinfo{person}{Rahul~G
  Krishnan}.} \bibinfo{year}{2022}\natexlab{}.
\newblock \showarticletitle{Partial Identification of Treatment Effects with
  Implicit Generative Models}.
\newblock \bibinfo{journal}{\emph{Advances in Neural Information Processing
  Systems}}  \bibinfo{volume}{35} (\bibinfo{year}{2022}),
  \bibinfo{pages}{22816--22829}.
\newblock


\bibitem[Berrevoets et~al\mbox{.}(2021)]%
        {berrevoets2021disentangled}
\bibfield{author}{\bibinfo{person}{Jeroen Berrevoets}, \bibinfo{person}{Alicia
  Curth}, \bibinfo{person}{Ioana Bica}, \bibinfo{person}{Eoin McKinney}, {and}
  \bibinfo{person}{Mihaela van~der Schaar}.} \bibinfo{year}{2021}\natexlab{}.
\newblock \showarticletitle{Disentangled counterfactual recurrent networks for
  treatment effect inference over time}.
\newblock \bibinfo{journal}{\emph{arXiv preprint arXiv:2112.03811}}
  (\bibinfo{year}{2021}).
\newblock


\bibitem[Bica et~al\mbox{.}(2019)]%
        {bica2019estimating}
\bibfield{author}{\bibinfo{person}{Ioana Bica}, \bibinfo{person}{Ahmed~M Alaa},
  \bibinfo{person}{James Jordon}, {and} \bibinfo{person}{Mihaela van~der
  Schaar}.} \bibinfo{year}{2019}\natexlab{}.
\newblock \showarticletitle{Estimating counterfactual treatment outcomes over
  time through adversarially balanced representations}. In
  \bibinfo{booktitle}{\emph{International Conference on Learning
  Representations}}.
\newblock


\bibitem[Bickel and Kwon(2001)]%
        {bickel2001inference}
\bibfield{author}{\bibinfo{person}{Peter~J Bickel} {and}
  \bibinfo{person}{Jaimyoung Kwon}.} \bibinfo{year}{2001}\natexlab{}.
\newblock \showarticletitle{Inference for semiparametric models: some questions
  and an answer}.
\newblock \bibinfo{journal}{\emph{Statistica Sinica}} (\bibinfo{year}{2001}),
  \bibinfo{pages}{863--886}.
\newblock


\bibitem[Bonvini et~al\mbox{.}(2021)]%
        {bonvini2021causal}
\bibfield{author}{\bibinfo{person}{Matteo Bonvini}, \bibinfo{person}{Edward
  Kennedy}, \bibinfo{person}{Valerie Ventura}, {and} \bibinfo{person}{Larry
  Wasserman}.} \bibinfo{year}{2021}\natexlab{}.
\newblock \showarticletitle{Causal inference in the time of Covid-19}.
\newblock \bibinfo{journal}{\emph{arXiv preprint arXiv:2103.04472}}
  (\bibinfo{year}{2021}).
\newblock


\bibitem[Bureau(2022)]%
        {uscensus}
\bibfield{author}{\bibinfo{person}{U.S.~Census Bureau}.}
  \bibinfo{year}{2022}\natexlab{}.
\newblock \bibinfo{title}{State Population Totals: 2020-2022}.
\newblock
  \bibinfo{howpublished}{\url{https://www.census.gov/data/tables/time-series/demo/popest/2020s-state-total.html}}.
\newblock
\newblock
\shownote{Accessed: 2022-09-15}.


\bibitem[Chen et~al\mbox{.}(2023)]%
        {chen2023multi}
\bibfield{author}{\bibinfo{person}{Yehu Chen}, \bibinfo{person}{Annamaria
  Prati}, \bibinfo{person}{Jacob Montgomery}, {and} \bibinfo{person}{Roman
  Garnett}.} \bibinfo{year}{2023}\natexlab{}.
\newblock \showarticletitle{A Multi-Task Gaussian Process Model for Inferring
  Time-Varying Treatment Effects in Panel Data}. In
  \bibinfo{booktitle}{\emph{International Conference on Artificial Intelligence
  and Statistics}}. PMLR, \bibinfo{pages}{4068--4088}.
\newblock


\bibitem[Chernozhukov et~al\mbox{.}(2013)]%
        {chernozhukov2013inference}
\bibfield{author}{\bibinfo{person}{Victor Chernozhukov},
  \bibinfo{person}{Iv{\'a}n Fern{\'a}ndez-Val}, {and} \bibinfo{person}{Blaise
  Melly}.} \bibinfo{year}{2013}\natexlab{}.
\newblock \showarticletitle{Inference on counterfactual distributions}.
\newblock \bibinfo{journal}{\emph{Econometrica}} \bibinfo{volume}{81},
  \bibinfo{number}{6} (\bibinfo{year}{2013}), \bibinfo{pages}{2205--2268}.
\newblock


\bibitem[Deng(2012)]%
        {deng2012mnist}
\bibfield{author}{\bibinfo{person}{Li Deng}.} \bibinfo{year}{2012}\natexlab{}.
\newblock \showarticletitle{The mnist database of handwritten digit images for
  machine learning research [best of the web]}.
\newblock \bibinfo{journal}{\emph{IEEE signal processing magazine}}
  \bibinfo{volume}{29}, \bibinfo{number}{6} (\bibinfo{year}{2012}),
  \bibinfo{pages}{141--142}.
\newblock


\bibitem[Dhariwal and Nichol(2021)]%
        {dhariwal2021diffusion}
\bibfield{author}{\bibinfo{person}{Prafulla Dhariwal} {and}
  \bibinfo{person}{Alexander Nichol}.} \bibinfo{year}{2021}\natexlab{}.
\newblock \showarticletitle{Diffusion models beat gans on image synthesis}.
\newblock \bibinfo{journal}{\emph{Advances in neural information processing
  systems}}  \bibinfo{volume}{34} (\bibinfo{year}{2021}),
  \bibinfo{pages}{8780--8794}.
\newblock


\bibitem[DiNardo et~al\mbox{.}(1996)]%
        {10.2307/2171954}
\bibfield{author}{\bibinfo{person}{John DiNardo}, \bibinfo{person}{Nicole~M.
  Fortin}, {and} \bibinfo{person}{Thomas Lemieux}.}
  \bibinfo{year}{1996}\natexlab{}.
\newblock \showarticletitle{Labor Market Institutions and the Distribution of
  Wages, 1973-1992: A Semiparametric Approach}.
\newblock \bibinfo{journal}{\emph{Econometrica}} \bibinfo{volume}{64},
  \bibinfo{number}{5} (\bibinfo{year}{1996}), \bibinfo{pages}{1001--1044}.
\newblock
\showISSN{00129682, 14680262}
\urldef\tempurl%
\url{http://www.jstor.org/stable/2171954}
\showURL{%
\tempurl}


\bibitem[Fitzmaurice et~al\mbox{.}(2008)]%
        {fitzmaurice2008longitudinal}
\bibfield{author}{\bibinfo{person}{Garrett Fitzmaurice}, \bibinfo{person}{Marie
  Davidian}, \bibinfo{person}{Geert Verbeke}, {and} \bibinfo{person}{Geert
  Molenberghs}.} \bibinfo{year}{2008}\natexlab{}.
\newblock \bibinfo{booktitle}{\emph{Longitudinal data analysis}}.
\newblock \bibinfo{publisher}{CRC press}.
\newblock


\bibitem[for Disease~Control(2021)]%
        {centers2021us}
\bibfield{author}{\bibinfo{person}{Centers for Disease~Control}.}
  \bibinfo{year}{2021}\natexlab{}.
\newblock \showarticletitle{US state and territorial public mask mandates from
  April 10, 2020 through August 15, 2021 by county by day}.
\newblock \bibinfo{journal}{\emph{Policy Surveillance. September}}
  \bibinfo{volume}{10} (\bibinfo{year}{2021}).
\newblock


\bibitem[Frauen et~al\mbox{.}(2023)]%
        {frauen2023estimating}
\bibfield{author}{\bibinfo{person}{Dennis Frauen}, \bibinfo{person}{Tobias
  Hatt}, \bibinfo{person}{Valentyn Melnychuk}, {and} \bibinfo{person}{Stefan
  Feuerriegel}.} \bibinfo{year}{2023}\natexlab{}.
\newblock \showarticletitle{Estimating average causal effects from patient
  trajectories}. In \bibinfo{booktitle}{\emph{Proceedings of the AAAI
  Conference on Artificial Intelligence}}. \bibinfo{pages}{7586--7594}.
\newblock


\bibitem[Frogner et~al\mbox{.}(2015)]%
        {frogner2015learning}
\bibfield{author}{\bibinfo{person}{Charlie Frogner}, \bibinfo{person}{Chiyuan
  Zhang}, \bibinfo{person}{Hossein Mobahi}, \bibinfo{person}{Mauricio Araya},
  {and} \bibinfo{person}{Tomaso~A Poggio}.} \bibinfo{year}{2015}\natexlab{}.
\newblock \showarticletitle{Learning with a Wasserstein loss}.
\newblock \bibinfo{journal}{\emph{Advances in neural information processing
  systems}}  \bibinfo{volume}{28} (\bibinfo{year}{2015}).
\newblock


\bibitem[Fujii et~al\mbox{.}(2022)]%
        {fujii2022estimating}
\bibfield{author}{\bibinfo{person}{Keisuke Fujii}, \bibinfo{person}{Koh
  Takeuchi}, \bibinfo{person}{Atsushi Kuribayashi}, \bibinfo{person}{Naoya
  Takeishi}, \bibinfo{person}{Yoshinobu Kawahara}, {and}
  \bibinfo{person}{Kazuya Takeda}.} \bibinfo{year}{2022}\natexlab{}.
\newblock \showarticletitle{Estimating counterfactual treatment outcomes over
  time in complex multi-agent scenarios}.
\newblock \bibinfo{journal}{\emph{arXiv preprint arXiv:2206.01900}}
  (\bibinfo{year}{2022}).
\newblock


\bibitem[Ganin et~al\mbox{.}(2016)]%
        {ganin2016domain}
\bibfield{author}{\bibinfo{person}{Yaroslav Ganin}, \bibinfo{person}{Evgeniya
  Ustinova}, \bibinfo{person}{Hana Ajakan}, \bibinfo{person}{Pascal Germain},
  \bibinfo{person}{Hugo Larochelle}, \bibinfo{person}{Fran{\c{c}}ois
  Laviolette}, \bibinfo{person}{Mario March}, {and} \bibinfo{person}{Victor
  Lempitsky}.} \bibinfo{year}{2016}\natexlab{}.
\newblock \showarticletitle{Domain-adversarial training of neural networks}.
\newblock \bibinfo{journal}{\emph{Journal of machine learning research}}
  \bibinfo{volume}{17}, \bibinfo{number}{59} (\bibinfo{year}{2016}),
  \bibinfo{pages}{1--35}.
\newblock


\bibitem[Goodfellow et~al\mbox{.}(2014)]%
        {goodfellow2014generative}
\bibfield{author}{\bibinfo{person}{Ian Goodfellow}, \bibinfo{person}{Jean
  Pouget-Abadie}, \bibinfo{person}{Mehdi Mirza}, \bibinfo{person}{Bing Xu},
  \bibinfo{person}{David Warde-Farley}, \bibinfo{person}{Sherjil Ozair},
  \bibinfo{person}{Aaron Courville}, {and} \bibinfo{person}{Yoshua Bengio}.}
  \bibinfo{year}{2014}\natexlab{}.
\newblock \showarticletitle{Generative adversarial nets}.
\newblock \bibinfo{journal}{\emph{Advances in neural information processing
  systems}}  \bibinfo{volume}{27} (\bibinfo{year}{2014}).
\newblock


\bibitem[Google(2022)]%
        {google}
\bibfield{author}{\bibinfo{person}{Google}.} \bibinfo{year}{2022}\natexlab{}.
\newblock \bibinfo{title}{Community Mobility Reports}.
\newblock
  \bibinfo{howpublished}{\url{https://www.google.com/covid19/mobility/}}.
\newblock
\newblock
\shownote{Accessed: 2022-09-15}.


\bibitem[Goudet et~al\mbox{.}(2017)]%
        {goudet2017causal}
\bibfield{author}{\bibinfo{person}{Olivier Goudet}, \bibinfo{person}{Diviyan
  Kalainathan}, \bibinfo{person}{Philippe Caillou}, \bibinfo{person}{Isabelle
  Guyon}, \bibinfo{person}{David Lopez-Paz}, {and} \bibinfo{person}{Mich{\`e}le
  Sebag}.} \bibinfo{year}{2017}\natexlab{}.
\newblock \showarticletitle{Causal generative neural networks}.
\newblock \bibinfo{journal}{\emph{arXiv preprint arXiv:1711.08936}}
  (\bibinfo{year}{2017}).
\newblock


\bibitem[Group(2022)]%
        {delphi}
\bibfield{author}{\bibinfo{person}{CMU~DELPHI Group}.}
  \bibinfo{year}{2022}\natexlab{}.
\newblock \bibinfo{title}{COVID-19 Symptom Surveys through Facebook}.
\newblock
  \bibinfo{howpublished}{\url{https://delphi.cmu.edu/blog/2020/08/26/covid-19-symptom-surveys-through-facebook/}}.
\newblock
\newblock
\shownote{Accessed: 2022-09-15}.


\bibitem[Guy~Jr et~al\mbox{.}(2021)]%
        {guy2021association}
\bibfield{author}{\bibinfo{person}{Gery~P Guy~Jr}, \bibinfo{person}{Florence~C
  Lee}, \bibinfo{person}{Gregory Sunshine}, \bibinfo{person}{Russell McCord},
  \bibinfo{person}{Mara Howard-Williams}, \bibinfo{person}{Lyudmyla
  Kompaniyets}, \bibinfo{person}{Christopher Dunphy}, \bibinfo{person}{Maxim
  Gakh}, \bibinfo{person}{Regen Weber}, \bibinfo{person}{Erin Sauber-Schatz},
  {et~al\mbox{.}}} \bibinfo{year}{2021}\natexlab{}.
\newblock \showarticletitle{Association of state-issued mask mandates and
  allowing on-premises restaurant dining with county-level COVID-19 case and
  death growth rates—United States, March 1--December 31, 2020}.
\newblock \bibinfo{journal}{\emph{Morbidity and Mortality Weekly Report}}
  \bibinfo{volume}{70}, \bibinfo{number}{10} (\bibinfo{year}{2021}),
  \bibinfo{pages}{350}.
\newblock


\bibitem[Hendrycks and Gimpel(2016)]%
        {hendrycks2016gaussian}
\bibfield{author}{\bibinfo{person}{Dan Hendrycks} {and} \bibinfo{person}{Kevin
  Gimpel}.} \bibinfo{year}{2016}\natexlab{}.
\newblock \showarticletitle{Gaussian error linear units (gelus)}.
\newblock \bibinfo{journal}{\emph{arXiv preprint arXiv:1606.08415}}
  (\bibinfo{year}{2016}).
\newblock


\bibitem[Heusel et~al\mbox{.}(2017)]%
        {heusel2017gans}
\bibfield{author}{\bibinfo{person}{Martin Heusel}, \bibinfo{person}{Hubert
  Ramsauer}, \bibinfo{person}{Thomas Unterthiner}, \bibinfo{person}{Bernhard
  Nessler}, {and} \bibinfo{person}{Sepp Hochreiter}.}
  \bibinfo{year}{2017}\natexlab{}.
\newblock \showarticletitle{Gans trained by a two time-scale update rule
  converge to a local nash equilibrium}.
\newblock \bibinfo{journal}{\emph{Advances in neural information processing
  systems}}  \bibinfo{volume}{30} (\bibinfo{year}{2017}).
\newblock


\bibitem[Hirano et~al\mbox{.}(2003)]%
        {hirano2003efficient}
\bibfield{author}{\bibinfo{person}{Keisuke Hirano}, \bibinfo{person}{Guido~W
  Imbens}, {and} \bibinfo{person}{Geert Ridder}.}
  \bibinfo{year}{2003}\natexlab{}.
\newblock \showarticletitle{Efficient estimation of average treatment effects
  using the estimated propensity score}.
\newblock \bibinfo{journal}{\emph{Econometrica}} \bibinfo{volume}{71},
  \bibinfo{number}{4} (\bibinfo{year}{2003}), \bibinfo{pages}{1161--1189}.
\newblock


\bibitem[Ho et~al\mbox{.}(2020)]%
        {ho2020denoising}
\bibfield{author}{\bibinfo{person}{Jonathan Ho}, \bibinfo{person}{Ajay Jain},
  {and} \bibinfo{person}{Pieter Abbeel}.} \bibinfo{year}{2020}\natexlab{}.
\newblock \showarticletitle{Denoising diffusion probabilistic models}.
\newblock \bibinfo{journal}{\emph{Advances in Neural Information Processing
  Systems}}  \bibinfo{volume}{33} (\bibinfo{year}{2020}),
  \bibinfo{pages}{6840--6851}.
\newblock


\bibitem[Ho and Salimans(2022)]%
        {ho2022classifier}
\bibfield{author}{\bibinfo{person}{Jonathan Ho} {and} \bibinfo{person}{Tim
  Salimans}.} \bibinfo{year}{2022}\natexlab{}.
\newblock \showarticletitle{Classifier-free diffusion guidance}.
\newblock \bibinfo{journal}{\emph{arXiv preprint arXiv:2207.12598}}
  (\bibinfo{year}{2022}).
\newblock


\bibitem[Im et~al\mbox{.}(2021)]%
        {im2021causal}
\bibfield{author}{\bibinfo{person}{Daniel~Jiwoong Im},
  \bibinfo{person}{Kyunghyun Cho}, {and} \bibinfo{person}{Narges Razavian}.}
  \bibinfo{year}{2021}\natexlab{}.
\newblock \showarticletitle{Causal effect variational autoencoder with uniform
  treatment}.
\newblock \bibinfo{journal}{\emph{arXiv preprint arXiv:2111.08656}}
  (\bibinfo{year}{2021}).
\newblock


\bibitem[Imai and Van~Dyk(2004)]%
        {imai2004causal}
\bibfield{author}{\bibinfo{person}{Kosuke Imai} {and} \bibinfo{person}{David~A
  Van~Dyk}.} \bibinfo{year}{2004}\natexlab{}.
\newblock \showarticletitle{Causal inference with general treatment regimes:
  Generalizing the propensity score}.
\newblock \bibinfo{journal}{\emph{J. Amer. Statist. Assoc.}}
  \bibinfo{volume}{99}, \bibinfo{number}{467} (\bibinfo{year}{2004}),
  \bibinfo{pages}{854--866}.
\newblock


\bibitem[Imbens(2004)]%
        {imbens2004nonparametric}
\bibfield{author}{\bibinfo{person}{Guido~W Imbens}.}
  \bibinfo{year}{2004}\natexlab{}.
\newblock \showarticletitle{Nonparametric estimation of average treatment
  effects under exogeneity: A review}.
\newblock \bibinfo{journal}{\emph{Review of Economics and statistics}}
  \bibinfo{volume}{86}, \bibinfo{number}{1} (\bibinfo{year}{2004}),
  \bibinfo{pages}{4--29}.
\newblock


\bibitem[Jesson et~al\mbox{.}(2021)]%
        {jesson2021quantifying}
\bibfield{author}{\bibinfo{person}{Andrew Jesson}, \bibinfo{person}{S{\"o}ren
  Mindermann}, \bibinfo{person}{Yarin Gal}, {and} \bibinfo{person}{Uri
  Shalit}.} \bibinfo{year}{2021}\natexlab{}.
\newblock \showarticletitle{Quantifying ignorance in individual-level
  causal-effect estimates under hidden confounding}. In
  \bibinfo{booktitle}{\emph{International Conference on Machine Learning}}.
  PMLR, \bibinfo{pages}{4829--4838}.
\newblock


\bibitem[Kennedy et~al\mbox{.}(2023)]%
        {Kennedy23}
\bibfield{author}{\bibinfo{person}{E~H Kennedy}, \bibinfo{person}{S
  Balakrishnan}, {and} \bibinfo{person}{L~A Wasserman}.}
  \bibinfo{year}{2023}\natexlab{}.
\newblock \showarticletitle{{Semiparametric counterfactual density
  estimation}}.
\newblock \bibinfo{journal}{\emph{Biometrika}} (\bibinfo{date}{03}
  \bibinfo{year}{2023}), \bibinfo{pages}{asad017}.
\newblock
\showISSN{1464-3510}
\urldef\tempurl%
\url{https://doi.org/10.1093/biomet/asad017}
\showDOI{\tempurl}
\showeprint{https://academic.oup.com/biomet/advance-article-pdf/doi/10.1093/biomet/asad017/50607309/asad017.pdf}


\bibitem[Kim et~al\mbox{.}(2018)]%
        {kim2018causal}
\bibfield{author}{\bibinfo{person}{Kwangho Kim}, \bibinfo{person}{Jisu Kim},
  {and} \bibinfo{person}{Edward~H Kennedy}.} \bibinfo{year}{2018}\natexlab{}.
\newblock \showarticletitle{Causal effects based on distributional distances}.
\newblock \bibinfo{journal}{\emph{arXiv preprint arXiv:1806.02935}}
  (\bibinfo{year}{2018}).
\newblock


\bibitem[Kingma and Ba(2014)]%
        {kingma2014adam}
\bibfield{author}{\bibinfo{person}{Diederik~P Kingma} {and}
  \bibinfo{person}{Jimmy Ba}.} \bibinfo{year}{2014}\natexlab{}.
\newblock \showarticletitle{Adam: A method for stochastic optimization}.
\newblock \bibinfo{journal}{\emph{arXiv preprint arXiv:1412.6980}}
  (\bibinfo{year}{2014}).
\newblock


\bibitem[Kingma and Welling(2013)]%
        {kingma2013auto}
\bibfield{author}{\bibinfo{person}{Diederik~P Kingma} {and}
  \bibinfo{person}{Max Welling}.} \bibinfo{year}{2013}\natexlab{}.
\newblock \showarticletitle{Auto-encoding variational bayes}.
\newblock \bibinfo{journal}{\emph{arXiv preprint arXiv:1312.6114}}
  (\bibinfo{year}{2013}).
\newblock


\bibitem[Kleinberg and Hripcsak(2011)]%
        {kleinberg2011review}
\bibfield{author}{\bibinfo{person}{Samantha Kleinberg} {and}
  \bibinfo{person}{George Hripcsak}.} \bibinfo{year}{2011}\natexlab{}.
\newblock \showarticletitle{A review of causal inference for biomedical
  informatics}.
\newblock \bibinfo{journal}{\emph{Journal of biomedical informatics}}
  \bibinfo{volume}{44}, \bibinfo{number}{6} (\bibinfo{year}{2011}),
  \bibinfo{pages}{1102--1112}.
\newblock


\bibitem[Kuzmanovic et~al\mbox{.}(2021)]%
        {kuzmanovic2021deconfounding}
\bibfield{author}{\bibinfo{person}{Milan Kuzmanovic}, \bibinfo{person}{Tobias
  Hatt}, {and} \bibinfo{person}{Stefan Feuerriegel}.}
  \bibinfo{year}{2021}\natexlab{}.
\newblock \showarticletitle{Deconfounding Temporal Autoencoder: estimating
  treatment effects over time using noisy proxies}. In
  \bibinfo{booktitle}{\emph{Machine Learning for Health}}. PMLR,
  \bibinfo{pages}{143--155}.
\newblock


\bibitem[Li et~al\mbox{.}(2021)]%
        {li2021g}
\bibfield{author}{\bibinfo{person}{Rui Li}, \bibinfo{person}{Stephanie Hu},
  \bibinfo{person}{Mingyu Lu}, \bibinfo{person}{Yuria Utsumi},
  \bibinfo{person}{Prithwish Chakraborty}, \bibinfo{person}{Daby~M Sow},
  \bibinfo{person}{Piyush Madan}, \bibinfo{person}{Jun Li},
  \bibinfo{person}{Mohamed Ghalwash}, \bibinfo{person}{Zach Shahn},
  {et~al\mbox{.}}} \bibinfo{year}{2021}\natexlab{}.
\newblock \showarticletitle{G-net: a recurrent network approach to
  g-computation for counterfactual prediction under a dynamic treatment
  regime}. In \bibinfo{booktitle}{\emph{Machine Learning for Health}}. PMLR,
  \bibinfo{pages}{282--299}.
\newblock


\bibitem[Lim et~al\mbox{.}(2018a)]%
        {lim2018forecasting}
\bibfield{author}{\bibinfo{person}{Bryan Lim}, \bibinfo{person}{Ahmed Alaa},
  {and} \bibinfo{person}{Mihaela van~der Schaar}.}
  \bibinfo{year}{2018}\natexlab{a}.
\newblock \showarticletitle{Forecasting Treatment Responses Over Time Using
  Recurrent Marginal Structural Networks}. In
  \bibinfo{booktitle}{\emph{Advances in Neural Information Processing
  Systems}}, \bibfield{editor}{\bibinfo{person}{S.~Bengio},
  \bibinfo{person}{H.~Wallach}, \bibinfo{person}{H.~Larochelle},
  \bibinfo{person}{K.~Grauman}, \bibinfo{person}{N.~Cesa-Bianchi}, {and}
  \bibinfo{person}{R.~Garnett}} (Eds.), Vol.~\bibinfo{volume}{31}.
  \bibinfo{publisher}{Curran Associates, Inc.}
\newblock
\urldef\tempurl%
\url{https://proceedings.neurips.cc/paper_files/paper/2018/file/56e6a93212e4482d99c84a639d254b67-Paper.pdf}
\showURL{%
\tempurl}


\bibitem[Lim et~al\mbox{.}(2018b)]%
        {lim2018molecular}
\bibfield{author}{\bibinfo{person}{Jaechang Lim}, \bibinfo{person}{Seongok
  Ryu}, \bibinfo{person}{Jin~Woo Kim}, {and} \bibinfo{person}{Woo~Youn Kim}.}
  \bibinfo{year}{2018}\natexlab{b}.
\newblock \showarticletitle{Molecular generative model based on conditional
  variational autoencoder for de novo molecular design}.
\newblock \bibinfo{journal}{\emph{Journal of cheminformatics}}
  \bibinfo{volume}{10}, \bibinfo{number}{1} (\bibinfo{year}{2018}),
  \bibinfo{pages}{1--9}.
\newblock


\bibitem[Liu et~al\mbox{.}(2022)]%
        {liu2022causalegm}
\bibfield{author}{\bibinfo{person}{Qiao Liu}, \bibinfo{person}{Zhongren Chen},
  {and} \bibinfo{person}{Wing~Hung Wong}.} \bibinfo{year}{2022}\natexlab{}.
\newblock \showarticletitle{CausalEGM: a general causal inference framework by
  encoding generative modeling}.
\newblock \bibinfo{journal}{\emph{arXiv preprint arXiv:2212.05925}}
  (\bibinfo{year}{2022}).
\newblock


\bibitem[Louizos et~al\mbox{.}(2017)]%
        {louizos2017causal}
\bibfield{author}{\bibinfo{person}{Christos Louizos}, \bibinfo{person}{Uri
  Shalit}, \bibinfo{person}{Joris~M Mooij}, \bibinfo{person}{David Sontag},
  \bibinfo{person}{Richard Zemel}, {and} \bibinfo{person}{Max Welling}.}
  \bibinfo{year}{2017}\natexlab{}.
\newblock \showarticletitle{Causal effect inference with deep latent-variable
  models}.
\newblock \bibinfo{journal}{\emph{Advances in neural information processing
  systems}}  \bibinfo{volume}{30} (\bibinfo{year}{2017}).
\newblock


\bibitem[Maltezou et~al\mbox{.}(2021)]%
        {maltezou2021post}
\bibfield{author}{\bibinfo{person}{Helena~C Maltezou},
  \bibinfo{person}{Androula Pavli}, {and} \bibinfo{person}{Athanasios
  Tsakris}.} \bibinfo{year}{2021}\natexlab{}.
\newblock \showarticletitle{Post-COVID syndrome: an insight on its
  pathogenesis}.
\newblock \bibinfo{journal}{\emph{Vaccines}} \bibinfo{volume}{9},
  \bibinfo{number}{5} (\bibinfo{year}{2021}), \bibinfo{pages}{497}.
\newblock


\bibitem[Melnychuk et~al\mbox{.}(2022)]%
        {melnychuk2022causal}
\bibfield{author}{\bibinfo{person}{Valentyn Melnychuk}, \bibinfo{person}{Dennis
  Frauen}, {and} \bibinfo{person}{Stefan Feuerriegel}.}
  \bibinfo{year}{2022}\natexlab{}.
\newblock \showarticletitle{Causal transformer for estimating counterfactual
  outcomes}. In \bibinfo{booktitle}{\emph{International Conference on Machine
  Learning}}. PMLR, \bibinfo{pages}{15293--15329}.
\newblock


\bibitem[Melnychuk et~al\mbox{.}(2023)]%
        {melnychuk2023normalizing}
\bibfield{author}{\bibinfo{person}{Valentyn Melnychuk}, \bibinfo{person}{Dennis
  Frauen}, {and} \bibinfo{person}{Stefan Feuerriegel}.}
  \bibinfo{year}{2023}\natexlab{}.
\newblock \showarticletitle{Normalizing flows for interventional density
  estimation}. In \bibinfo{booktitle}{\emph{International Conference on Machine
  Learning}}. PMLR, \bibinfo{pages}{24361--24397}.
\newblock


\bibitem[Mirza and Osindero(2014)]%
        {mirza2014conditional}
\bibfield{author}{\bibinfo{person}{Mehdi Mirza} {and} \bibinfo{person}{Simon
  Osindero}.} \bibinfo{year}{2014}\natexlab{}.
\newblock \bibinfo{title}{Conditional Generative Adversarial Nets}.
\newblock
\newblock
\urldef\tempurl%
\url{http://arxiv.org/abs/1411.1784}
\showURL{%
\tempurl}
\newblock
\shownote{cite arxiv:1411.1784}.


\bibitem[Mishra et~al\mbox{.}(2018)]%
        {mishra2018generative}
\bibfield{author}{\bibinfo{person}{Ashish Mishra}, \bibinfo{person}{Shiva
  Krishna~Reddy}, \bibinfo{person}{Anurag Mittal}, {and}
  \bibinfo{person}{Hema~A Murthy}.} \bibinfo{year}{2018}\natexlab{}.
\newblock \showarticletitle{A generative model for zero shot learning using
  conditional variational autoencoders}. In
  \bibinfo{booktitle}{\emph{Proceedings of the IEEE conference on computer
  vision and pattern recognition workshops}}. \bibinfo{pages}{2188--2196}.
\newblock


\bibitem[Murphy(2012)]%
        {murphy2012machine}
\bibfield{author}{\bibinfo{person}{Kevin~P Murphy}.}
  \bibinfo{year}{2012}\natexlab{}.
\newblock \bibinfo{booktitle}{\emph{Machine learning: a probabilistic
  perspective}}.
\newblock \bibinfo{publisher}{MIT press}.
\newblock


\bibitem[Neyman(1923)]%
        {neyman1923applications}
\bibfield{author}{\bibinfo{person}{Jersey Neyman}.}
  \bibinfo{year}{1923}\natexlab{}.
\newblock \showarticletitle{Sur les applications de la th{\'e}orie des
  probabilit{\'e}s aux experiences agricoles: Essai des principes}.
\newblock \bibinfo{journal}{\emph{Roczniki Nauk Rolniczych}}
  \bibinfo{volume}{10}, \bibinfo{number}{1} (\bibinfo{year}{1923}),
  \bibinfo{pages}{1--51}.
\newblock


\bibitem[Nguyen(2021)]%
        {nguyen2021mask}
\bibfield{author}{\bibinfo{person}{My Nguyen}.}
  \bibinfo{year}{2021}\natexlab{}.
\newblock \showarticletitle{Mask mandates and COVID-19 related symptoms in the
  US}.
\newblock \bibinfo{journal}{\emph{ClinicoEconomics and Outcomes Research}}
  (\bibinfo{year}{2021}), \bibinfo{pages}{757--766}.
\newblock


\bibitem[Pagnoni et~al\mbox{.}(2018)]%
        {pagnoni2018conditional}
\bibfield{author}{\bibinfo{person}{Artidoro Pagnoni}, \bibinfo{person}{Kevin
  Liu}, {and} \bibinfo{person}{Shangyan Li}.} \bibinfo{year}{2018}\natexlab{}.
\newblock \showarticletitle{Conditional variational autoencoder for neural
  machine translation}.
\newblock \bibinfo{journal}{\emph{arXiv preprint arXiv:1812.04405}}
  (\bibinfo{year}{2018}).
\newblock


\bibitem[Panaretos and Zemel(2019)]%
        {panaretos2019statistical}
\bibfield{author}{\bibinfo{person}{Victor~M Panaretos} {and}
  \bibinfo{person}{Yoav Zemel}.} \bibinfo{year}{2019}\natexlab{}.
\newblock \showarticletitle{Statistical aspects of Wasserstein distances}.
\newblock \bibinfo{journal}{\emph{Annual review of statistics and its
  application}}  \bibinfo{volume}{6} (\bibinfo{year}{2019}),
  \bibinfo{pages}{405--431}.
\newblock


\bibitem[Pearl(2009)]%
        {pearl2009causal}
\bibfield{author}{\bibinfo{person}{Judea Pearl}.}
  \bibinfo{year}{2009}\natexlab{}.
\newblock \showarticletitle{Causal inference in statistics: An overview}.
\newblock  (\bibinfo{year}{2009}).
\newblock


\bibitem[Reynaud et~al\mbox{.}(2022)]%
        {reynaud2022d}
\bibfield{author}{\bibinfo{person}{Hadrien Reynaud},
  \bibinfo{person}{Athanasios Vlontzos}, \bibinfo{person}{Mischa Dombrowski},
  \bibinfo{person}{Ciar{\'a}n Gilligan~Lee}, \bibinfo{person}{Arian Beqiri},
  \bibinfo{person}{Paul Leeson}, {and} \bibinfo{person}{Bernhard Kainz}.}
  \bibinfo{year}{2022}\natexlab{}.
\newblock \showarticletitle{D’artagnan: Counterfactual video generation}. In
  \bibinfo{booktitle}{\emph{Medical Image Computing and Computer Assisted
  Intervention--MICCAI 2022: 25th International Conference, Singapore,
  September 18--22, 2022, Proceedings, Part VIII}}. Springer,
  \bibinfo{pages}{599--609}.
\newblock


\bibitem[Robins(1986)]%
        {robins1986new}
\bibfield{author}{\bibinfo{person}{James Robins}.}
  \bibinfo{year}{1986}\natexlab{}.
\newblock \showarticletitle{A new approach to causal inference in mortality
  studies with a sustained exposure period—application to control of the
  healthy worker survivor effect}.
\newblock \bibinfo{journal}{\emph{Mathematical modelling}} \bibinfo{volume}{7},
  \bibinfo{number}{9-12} (\bibinfo{year}{1986}), \bibinfo{pages}{1393--1512}.
\newblock


\bibitem[Robins and Hernan(2008)]%
        {robins2008estimation}
\bibfield{author}{\bibinfo{person}{James Robins} {and} \bibinfo{person}{Miguel
  Hernan}.} \bibinfo{year}{2008}\natexlab{}.
\newblock \showarticletitle{Estimation of the causal effects of time-varying
  exposures}.
\newblock \bibinfo{journal}{\emph{Chapman \& Hall/CRC Handbooks of Modern
  Statistical Methods}} (\bibinfo{year}{2008}), \bibinfo{pages}{553--599}.
\newblock


\bibitem[Robins(1994)]%
        {robins1994correcting}
\bibfield{author}{\bibinfo{person}{James~M Robins}.}
  \bibinfo{year}{1994}\natexlab{}.
\newblock \showarticletitle{Correcting for non-compliance in randomized trials
  using structural nested mean models}.
\newblock \bibinfo{journal}{\emph{Communications in Statistics-Theory and
  methods}} \bibinfo{volume}{23}, \bibinfo{number}{8} (\bibinfo{year}{1994}),
  \bibinfo{pages}{2379--2412}.
\newblock


\bibitem[Robins(1999)]%
        {robins1999association}
\bibfield{author}{\bibinfo{person}{James~M Robins}.}
  \bibinfo{year}{1999}\natexlab{}.
\newblock \showarticletitle{Association, causation, and marginal structural
  models}.
\newblock \bibinfo{journal}{\emph{Synthese}} \bibinfo{volume}{121},
  \bibinfo{number}{1/2} (\bibinfo{year}{1999}), \bibinfo{pages}{151--179}.
\newblock


\bibitem[Robins et~al\mbox{.}(1999)]%
        {robins1999estimation}
\bibfield{author}{\bibinfo{person}{James~M Robins}, \bibinfo{person}{Sander
  Greenland}, {and} \bibinfo{person}{Fu-Chang Hu}.}
  \bibinfo{year}{1999}\natexlab{}.
\newblock \showarticletitle{Estimation of the causal effect of a time-varying
  exposure on the marginal mean of a repeated binary outcome}.
\newblock \bibinfo{journal}{\emph{J. Amer. Statist. Assoc.}}
  \bibinfo{volume}{94}, \bibinfo{number}{447} (\bibinfo{year}{1999}),
  \bibinfo{pages}{687--700}.
\newblock


\bibitem[Robins et~al\mbox{.}(2000)]%
        {robins2000marginal}
\bibfield{author}{\bibinfo{person}{James~M Robins},
  \bibinfo{person}{Miguel~Angel Hernan}, {and} \bibinfo{person}{Babette
  Brumback}.} \bibinfo{year}{2000}\natexlab{}.
\newblock \showarticletitle{Marginal structural models and causal inference in
  epidemiology}.
\newblock \bibinfo{journal}{\emph{Epidemiology}} (\bibinfo{year}{2000}),
  \bibinfo{pages}{550--560}.
\newblock


\bibitem[Robins et~al\mbox{.}(1994)]%
        {robins1994estimation}
\bibfield{author}{\bibinfo{person}{James~M Robins}, \bibinfo{person}{Andrea
  Rotnitzky}, {and} \bibinfo{person}{Lue~Ping Zhao}.}
  \bibinfo{year}{1994}\natexlab{}.
\newblock \showarticletitle{Estimation of regression coefficients when some
  regressors are not always observed}.
\newblock \bibinfo{journal}{\emph{Journal of the American statistical
  Association}} \bibinfo{volume}{89}, \bibinfo{number}{427}
  (\bibinfo{year}{1994}), \bibinfo{pages}{846--866}.
\newblock


\bibitem[Rombach et~al\mbox{.}(2022)]%
        {rombach2022high}
\bibfield{author}{\bibinfo{person}{Robin Rombach}, \bibinfo{person}{Andreas
  Blattmann}, \bibinfo{person}{Dominik Lorenz}, \bibinfo{person}{Patrick
  Esser}, {and} \bibinfo{person}{Bj{\"o}rn Ommer}.}
  \bibinfo{year}{2022}\natexlab{}.
\newblock \showarticletitle{High-resolution image synthesis with latent
  diffusion models}. In \bibinfo{booktitle}{\emph{Proceedings of the IEEE/CVF
  conference on computer vision and pattern recognition}}.
  \bibinfo{pages}{10684--10695}.
\newblock


\bibitem[Ronneberger et~al\mbox{.}(2015)]%
        {ronneberger2015u}
\bibfield{author}{\bibinfo{person}{Olaf Ronneberger}, \bibinfo{person}{Philipp
  Fischer}, {and} \bibinfo{person}{Thomas Brox}.}
  \bibinfo{year}{2015}\natexlab{}.
\newblock \showarticletitle{U-net: Convolutional networks for biomedical image
  segmentation}. In \bibinfo{booktitle}{\emph{Medical Image Computing and
  Computer-Assisted Intervention--MICCAI 2015: 18th International Conference,
  Munich, Germany, October 5-9, 2015, Proceedings, Part III 18}}. Springer,
  \bibinfo{pages}{234--241}.
\newblock


\bibitem[Rosenbaum and Rubin(1983)]%
        {rosenbaum1983central}
\bibfield{author}{\bibinfo{person}{Paul~R Rosenbaum} {and}
  \bibinfo{person}{Donald~B Rubin}.} \bibinfo{year}{1983}\natexlab{}.
\newblock \showarticletitle{The central role of the propensity score in
  observational studies for causal effects}.
\newblock \bibinfo{journal}{\emph{Biometrika}} \bibinfo{volume}{70},
  \bibinfo{number}{1} (\bibinfo{year}{1983}), \bibinfo{pages}{41--55}.
\newblock


\bibitem[Rosenblatt(1956)]%
        {rosenblatt1956remarks}
\bibfield{author}{\bibinfo{person}{Murray Rosenblatt}.}
  \bibinfo{year}{1956}\natexlab{}.
\newblock \showarticletitle{Remarks on some nonparametric estimates of a
  density function}.
\newblock \bibinfo{journal}{\emph{The annals of mathematical statistics}}
  (\bibinfo{year}{1956}), \bibinfo{pages}{832--837}.
\newblock


\bibitem[Rubin(1978)]%
        {rubin1978bayesian}
\bibfield{author}{\bibinfo{person}{Donald~B Rubin}.}
  \bibinfo{year}{1978}\natexlab{}.
\newblock \showarticletitle{Bayesian inference for causal effects: The role of
  randomization}.
\newblock \bibinfo{journal}{\emph{The Annals of statistics}}
  (\bibinfo{year}{1978}), \bibinfo{pages}{34--58}.
\newblock


\bibitem[Saini et~al\mbox{.}(2019)]%
        {saini2019multiple}
\bibfield{author}{\bibinfo{person}{Shiv~Kumar Saini}, \bibinfo{person}{Sunny
  Dhamnani}, \bibinfo{person}{Akil~Arif Ibrahim}, {and}
  \bibinfo{person}{Prithviraj Chavan}.} \bibinfo{year}{2019}\natexlab{}.
\newblock \showarticletitle{Multiple treatment effect estimation using deep
  generative model with task embedding}. In \bibinfo{booktitle}{\emph{The World
  Wide Web Conference}}. \bibinfo{pages}{1601--1611}.
\newblock


\bibitem[Sauer and Geiger(2021)]%
        {sauer2021counterfactual}
\bibfield{author}{\bibinfo{person}{Axel Sauer} {and} \bibinfo{person}{Andreas
  Geiger}.} \bibinfo{year}{2021}\natexlab{}.
\newblock \showarticletitle{Counterfactual generative networks}.
\newblock \bibinfo{journal}{\emph{arXiv preprint arXiv:2101.06046}}
  (\bibinfo{year}{2021}).
\newblock


\bibitem[Schulam and Saria(2017)]%
        {schulam2017reliable}
\bibfield{author}{\bibinfo{person}{Peter Schulam} {and} \bibinfo{person}{Suchi
  Saria}.} \bibinfo{year}{2017}\natexlab{}.
\newblock \showarticletitle{Reliable decision support using counterfactual
  models}.
\newblock \bibinfo{journal}{\emph{Advances in neural information processing
  systems}}  \bibinfo{volume}{30} (\bibinfo{year}{2017}).
\newblock


\bibitem[Scott and Thompson(1983)]%
        {scott1983probability}
\bibfield{author}{\bibinfo{person}{David~W Scott} {and}
  \bibinfo{person}{James~R Thompson}.} \bibinfo{year}{1983}\natexlab{}.
\newblock \showarticletitle{Probability density estimation in higher
  dimensions}. In \bibinfo{booktitle}{\emph{Computer Science and Statistics:
  Proceedings of the fifteenth symposium on the interface}},
  Vol.~\bibinfo{volume}{528}. North-Holland, Amsterdam,
  \bibinfo{pages}{173--179}.
\newblock


\bibitem[Seedat et~al\mbox{.}(2022)]%
        {seedat2022continuous}
\bibfield{author}{\bibinfo{person}{Nabeel Seedat}, \bibinfo{person}{Fergus
  Imrie}, \bibinfo{person}{Alexis Bellot}, \bibinfo{person}{Zhaozhi Qian},
  {and} \bibinfo{person}{Mihaela van~der Schaar}.}
  \bibinfo{year}{2022}\natexlab{}.
\newblock \showarticletitle{Continuous-Time Modeling of Counterfactual Outcomes
  Using Neural Controlled Differential Equations}.
\newblock \bibinfo{journal}{\emph{arXiv preprint arXiv:2206.08311}}
  (\bibinfo{year}{2022}).
\newblock


\bibitem[Sohl-Dickstein et~al\mbox{.}(2015a)]%
        {sohl2015deep}
\bibfield{author}{\bibinfo{person}{Jascha Sohl-Dickstein},
  \bibinfo{person}{Eric Weiss}, \bibinfo{person}{Niru Maheswaranathan}, {and}
  \bibinfo{person}{Surya Ganguli}.} \bibinfo{year}{2015}\natexlab{a}.
\newblock \showarticletitle{Deep unsupervised learning using nonequilibrium
  thermodynamics}. In \bibinfo{booktitle}{\emph{International conference on
  machine learning}}. PMLR, \bibinfo{pages}{2256--2265}.
\newblock


\bibitem[Sohl-Dickstein et~al\mbox{.}(2015b)]%
        {sohl-dickstein2015deep}
\bibfield{author}{\bibinfo{person}{Jascha Sohl-Dickstein},
  \bibinfo{person}{Eric Weiss}, \bibinfo{person}{Niru Maheswaranathan}, {and}
  \bibinfo{person}{Surya Ganguli}.} \bibinfo{year}{2015}\natexlab{b}.
\newblock \showarticletitle{Deep Unsupervised Learning using Nonequilibrium
  Thermodynamics}. In \bibinfo{booktitle}{\emph{Proceedings of the 32nd
  International Conference on Machine Learning}}
  \emph{(\bibinfo{series}{Proceedings of Machine Learning Research},
  Vol.~\bibinfo{volume}{37})}, \bibfield{editor}{\bibinfo{person}{Francis Bach}
  {and} \bibinfo{person}{David Blei}} (Eds.). \bibinfo{publisher}{PMLR},
  \bibinfo{address}{Lille, France}, \bibinfo{pages}{2256--2265}.
\newblock
\urldef\tempurl%
\url{https://proceedings.mlr.press/v37/sohl-dickstein15.html}
\showURL{%
\tempurl}


\bibitem[Sohn et~al\mbox{.}(2015)]%
        {sohn2015learning}
\bibfield{author}{\bibinfo{person}{Kihyuk Sohn}, \bibinfo{person}{Honglak Lee},
  {and} \bibinfo{person}{Xinchen Yan}.} \bibinfo{year}{2015}\natexlab{}.
\newblock \showarticletitle{Learning structured output representation using
  deep conditional generative models}.
\newblock \bibinfo{journal}{\emph{Advances in neural information processing
  systems}}  \bibinfo{volume}{28} (\bibinfo{year}{2015}).
\newblock


\bibitem[Times(2021)]%
        {nyt}
\bibfield{author}{\bibinfo{person}{The New~York Times}.}
  \bibinfo{year}{2021}\natexlab{}.
\newblock \bibinfo{title}{Coronavirus (Covid-19) Data in the United States}.
\newblock
  \bibinfo{howpublished}{\url{https://github.com/nytimes/covid-19-data}}.
\newblock
\newblock
\shownote{Accessed: 2022-09-15}.


\bibitem[Van~Dyke et~al\mbox{.}(2020)]%
        {van2020trends}
\bibfield{author}{\bibinfo{person}{Miriam~E Van~Dyke}, \bibinfo{person}{Tia~M
  Rogers}, \bibinfo{person}{Eric Pevzner}, \bibinfo{person}{Catherine~L
  Satterwhite}, \bibinfo{person}{Hina~B Shah}, \bibinfo{person}{Wyatt~J
  Beckman}, \bibinfo{person}{Farah Ahmed}, \bibinfo{person}{D~Charles Hunt},
  {and} \bibinfo{person}{John Rule}.} \bibinfo{year}{2020}\natexlab{}.
\newblock \showarticletitle{Trends in county-level COVID-19 incidence in
  counties with and without a mask mandate—Kansas, June 1--August 23, 2020}.
\newblock \bibinfo{journal}{\emph{Morbidity and Mortality Weekly Report}}
  \bibinfo{volume}{69}, \bibinfo{number}{47} (\bibinfo{year}{2020}),
  \bibinfo{pages}{1777}.
\newblock


\bibitem[Van~Looveren et~al\mbox{.}(2021)]%
        {van2021conditional}
\bibfield{author}{\bibinfo{person}{Arnaud Van~Looveren}, \bibinfo{person}{Janis
  Klaise}, \bibinfo{person}{Giovanni Vacanti}, {and} \bibinfo{person}{Oliver
  Cobb}.} \bibinfo{year}{2021}\natexlab{}.
\newblock \showarticletitle{Conditional generative models for counterfactual
  explanations}.
\newblock \bibinfo{journal}{\emph{arXiv preprint arXiv:2101.10123}}
  (\bibinfo{year}{2021}).
\newblock


\bibitem[Vanderschueren et~al\mbox{.}(2023)]%
        {vanderschueren2023accounting}
\bibfield{author}{\bibinfo{person}{Toon Vanderschueren},
  \bibinfo{person}{Alicia Curth}, \bibinfo{person}{Wouter Verbeke}, {and}
  \bibinfo{person}{Mihaela van~der Schaar}.} \bibinfo{year}{2023}\natexlab{}.
\newblock \showarticletitle{Accounting For Informative Sampling When Learning
  to Forecast Treatment Outcomes Over Time}.
\newblock \bibinfo{journal}{\emph{arXiv preprint arXiv:2306.04255}}
  (\bibinfo{year}{2023}).
\newblock


\bibitem[Wang et~al\mbox{.}(2018)]%
        {wang2018quantile}
\bibfield{author}{\bibinfo{person}{Lan Wang}, \bibinfo{person}{Yu Zhou},
  \bibinfo{person}{Rui Song}, {and} \bibinfo{person}{Ben Sherwood}.}
  \bibinfo{year}{2018}\natexlab{}.
\newblock \showarticletitle{Quantile-optimal treatment regimes}.
\newblock \bibinfo{journal}{\emph{J. Amer. Statist. Assoc.}}
  \bibinfo{volume}{113}, \bibinfo{number}{523} (\bibinfo{year}{2018}),
  \bibinfo{pages}{1243--1254}.
\newblock


\bibitem[Wang et~al\mbox{.}(2021)]%
        {wang2021effective}
\bibfield{author}{\bibinfo{person}{Yuxin Wang}, \bibinfo{person}{Zicheng Deng},
  {and} \bibinfo{person}{Donglu Shi}.} \bibinfo{year}{2021}\natexlab{}.
\newblock \showarticletitle{How effective is a mask in preventing COVID-19
  infection?}
\newblock \bibinfo{journal}{\emph{Medical devices \& sensors}}
  \bibinfo{volume}{4}, \bibinfo{number}{1} (\bibinfo{year}{2021}),
  \bibinfo{pages}{e10163}.
\newblock


\bibitem[Xiao et~al\mbox{.}(2010)]%
        {xiao2010accuracy}
\bibfield{author}{\bibinfo{person}{Yongling Xiao}, \bibinfo{person}{Michal
  Abrahamowicz}, {and} \bibinfo{person}{Erica~EM Moodie}.}
  \bibinfo{year}{2010}\natexlab{}.
\newblock \showarticletitle{Accuracy of conventional and marginal structural
  Cox model estimators: a simulation study}.
\newblock \bibinfo{journal}{\emph{The international journal of biostatistics}}
  \bibinfo{volume}{6}, \bibinfo{number}{2} (\bibinfo{year}{2010}).
\newblock


\bibitem[Yang et~al\mbox{.}(2023)]%
        {yang2023diffusion}
\bibfield{author}{\bibinfo{person}{Ling Yang}, \bibinfo{person}{Zhilong Zhang},
  \bibinfo{person}{Yang Song}, \bibinfo{person}{Shenda Hong},
  \bibinfo{person}{Runsheng Xu}, \bibinfo{person}{Yue Zhao},
  \bibinfo{person}{Wentao Zhang}, \bibinfo{person}{Bin Cui}, {and}
  \bibinfo{person}{Ming-Hsuan Yang}.} \bibinfo{year}{2023}\natexlab{}.
\newblock \showarticletitle{Diffusion models: A comprehensive survey of methods
  and applications}.
\newblock \bibinfo{journal}{\emph{Comput. Surveys}} \bibinfo{volume}{56},
  \bibinfo{number}{4} (\bibinfo{year}{2023}), \bibinfo{pages}{1--39}.
\newblock


\bibitem[Yoon et~al\mbox{.}(2018)]%
        {yoon2018ganite}
\bibfield{author}{\bibinfo{person}{Jinsung Yoon}, \bibinfo{person}{James
  Jordon}, {and} \bibinfo{person}{Mihaela Van Der~Schaar}.}
  \bibinfo{year}{2018}\natexlab{}.
\newblock \showarticletitle{GANITE: Estimation of individualized treatment
  effects using generative adversarial nets}. In
  \bibinfo{booktitle}{\emph{International conference on learning
  representations}}.
\newblock


\bibitem[Zhang et~al\mbox{.}(2017)]%
        {zhang2017mining}
\bibfield{author}{\bibinfo{person}{Weijia Zhang}, \bibinfo{person}{Thuc~Duy
  Le}, \bibinfo{person}{Lin Liu}, \bibinfo{person}{Zhi-Hua Zhou}, {and}
  \bibinfo{person}{Jiuyong Li}.} \bibinfo{year}{2017}\natexlab{}.
\newblock \showarticletitle{Mining heterogeneous causal effects for
  personalized cancer treatment}.
\newblock \bibinfo{journal}{\emph{Bioinformatics}} \bibinfo{volume}{33},
  \bibinfo{number}{15} (\bibinfo{year}{2017}), \bibinfo{pages}{2372--2378}.
\newblock


\bibitem[Zhang et~al\mbox{.}(2022)]%
        {zhang2022exploring}
\bibfield{author}{\bibinfo{person}{YiFan Zhang}, \bibinfo{person}{Hanlin
  Zhang}, \bibinfo{person}{Zachary~Chase Lipton}, \bibinfo{person}{Li~Erran
  Li}, {and} \bibinfo{person}{Eric Xing}.} \bibinfo{year}{2022}\natexlab{}.
\newblock \showarticletitle{Exploring transformer backbones for heterogeneous
  treatment effect estimation}. In \bibinfo{booktitle}{\emph{NeurIPS ML Safety
  Workshop}}.
\newblock


\bibitem[Zhu et~al\mbox{.}(2021)]%
        {zhu2021high}
\bibfield{author}{\bibinfo{person}{Shixiang Zhu}, \bibinfo{person}{Alexander
  Bukharin}, \bibinfo{person}{Liyan Xie}, \bibinfo{person}{Mauricio
  Santillana}, \bibinfo{person}{Shihao Yang}, {and} \bibinfo{person}{Yao Xie}.}
  \bibinfo{year}{2021}\natexlab{}.
\newblock \showarticletitle{High-Resolution Spatio-Temporal Model for
  County-Level COVID-19 Activity in the {U.S.}}
\newblock \bibinfo{journal}{\emph{ACM Trans. Manage. Inf. Syst.}}
  \bibinfo{volume}{12}, \bibinfo{number}{4}, Article \bibinfo{articleno}{33}
  (\bibinfo{date}{sep} \bibinfo{year}{2021}), \bibinfo{numpages}{20}~pages.
\newblock
\showISSN{2158-656X}
\urldef\tempurl%
\url{https://doi.org/10.1145/3468876}
\showDOI{\tempurl}


\bibitem[Zhu et~al\mbox{.}(2022)]%
        {zhu2022early}
\bibfield{author}{\bibinfo{person}{Shixiang Zhu}, \bibinfo{person}{Alexander
  Bukharin}, \bibinfo{person}{Liyan Xie}, \bibinfo{person}{Khurram Yamin},
  \bibinfo{person}{Shihao Yang}, \bibinfo{person}{Pinar Keskinocak}, {and}
  \bibinfo{person}{Yao Xie}.} \bibinfo{year}{2022}\natexlab{}.
\newblock \showarticletitle{Early Detection of {COVID-19} Hotspots Using
  Spatio-Temporal Data}.
\newblock \bibinfo{journal}{\emph{IEEE Journal of Selected Topics in Signal
  Processing}} \bibinfo{volume}{16}, \bibinfo{number}{2}
  (\bibinfo{year}{2022}), \bibinfo{pages}{250--260}.
\newblock
\urldef\tempurl%
\url{https://doi.org/10.1109/JSTSP.2022.3154972}
\showDOI{\tempurl}


\end{thebibliography}
\bibliographystyle{ACM-Reference-Format}

\newpage
\appendix

\section{Proof of Equation ~\ref{eq:log-likelihood}}\label{append:proof-obj}

The objective function for training the counterfactual generator minimizes the difference between $f_\theta(\cdot|\ab)$ and the true counterfactual distribution $f_{\ab}$ with respect to a distributional difference $D_f(\cdot,\cdot)$ over all treatment combinations, $\ab$. When the distance measure is the KL-divergence, Equation \ref{eq:obj-general} can be written as:
\begin{align*}
    \hat{\theta} 
    =&~ \argmin_{\theta \in \Theta}~\EE_{\AB} ~\left[{\rm KL}(f_{\ab}(\cdot) || f_\theta(\cdot|\ab)) \right] \\
    =&~ \argmin_{\theta \in \Theta}~\EE_{\AB} \left[\int \log \left(\frac{f_{\ab}(y)}{f_\theta(y|\ab)}\right)f_{\ab}(y) dy\right]\\
    =&~ \argmax_{\theta \in \Theta}~\EE_{\AB} \left[ \int \log \left(f_\theta(y|\ab)\right)f_{\ab}(y) dy \right]\\
    =&~ \argmax_{\theta \in \Theta}~\EE_{\AB}~\left[\EE_{y \sim f_{\ab}} ~\log f_\theta(\cdot|\ab)\right].  \\ 
\end{align*}

\section{Proof of Lemma~\ref{lemma:pseudo-prob}}
\label{append:proof-lemma-1}

Given a probability distribution for $(Y, \AB, \XB)$ and a causal directed acyclic graph (DAG) shown in Figure~\ref{fig:graphical-model}, we can factor $f(y, \ab, \xb)$ as:
\begin{equation}
    f(y, \ab, \xb) =\prob{y|\ab,\xb} \prod_{\tau=t-d+1}^{t} \Big[ \prob{x_{\tau}|\ab_{\tau-1},\xb_{\tau-1}}\cdot 
 \prob{a_{\tau}|\ab_{\tau-1},\xb_{\tau}}\Big].
    \label{eq:joint-prob-factor}
\end{equation}  
Using the definition of g-formula \citep{robins1999association}, we have:
\allowdisplaybreaks
\begin{align*}  
    \cprob{\ab}{y}
    =&~ \int \prob{y|\ab,\xb} \cdot \prod_{\tau=t-d+1}^{t} \prob{x_\tau|\ab_{\tau-1},\xb_{\tau-1}} d\xb\\
    =&~ \int \prob{y|\ab,\xb} \cdot \quad \frac{\prod_{\tau=t-d+1}^{t} \prob{a_{\tau}|\ab_{\tau-1},\xb_{\tau}}}{\prod_{\tau=t-d+1}^{t} \prob{a_{\tau}|\ab_{\tau-1},\xb_{\tau}}} \\
    & \cdot \prod_{\tau=t-d+1}^{t} \prob{ x_{\tau}|\ab_{\tau-1},\xb_{\tau-1}} d\xb \\
    \overset{(i)}{=}&~ \int \frac{1}{\prod_{\tau=t-d+1}^{t} \prob{a_{\tau}|\ab_{\tau-1},\xb_{\tau}}} \prob{y,\ab,\xb} d\xb,
\end{align*}
where the equation $(i)$ holds due to \eqref{eq:joint-prob-factor}.

\section{Proof of Proposition~\ref{prop:pesudo-objective}}

We recall our notations for densities: $f(y, \ab, \xb)$ denotes the density of the observed data, $f_{\ab}$ denotes the counterfactual density under $\ab$, and $f_\theta(\cdot|\ab)$ denotes the conditional density represented by our conditional generator. Note that these density notations should be interpreted in a broad sense to unify discrete and continuous random variables, meaning that when $\bar{a}$ is a discrete random variable, we allow the density function to be Delta functions. For example, when $\bar{A}$ is distributed as $\mathbb{P}(\bar{A} = \bar{a}_1) = \mathbb{P}(\bar{A} = \bar{a}_2) = 0.5$, its corresponding density function is $p(\cdot) = 0.5 \delta_{\bar{a}_1}(\cdot) + 0.5\delta_{\bar{a}_2}(\cdot)$.

We also recall that $w(\ab,\xb) = 1 / \prod_{\tau=t-d+1}^{t} \prob{a_{\tau}|\ab_{\tau-1},\xb_{\tau}}$ where $\prob{a_{\tau}|\ab_{\tau-1}, \xb_{\tau}}$ denotes the individual propensity score. Here $\ab$ is the collection of all the history treatment. Using Lemma~\ref{lemma:pseudo-prob}, we have:
\allowdisplaybreaks
\begin{align*}
& \quad \EE_{\AB}~\left[\EE_{y \sim f_{\ab}} [\log f_\theta(y|\ab) ]\right] && \\
& = \EE_{\AB}~\left[ \int \log f_\theta(y|\ab) \cprob{\ab}{y} dy \right] \\
& = \EE_{\AB}~\left[\int \log f_\theta(y|\ab) \int w(\ab,\xb) \prob{y,\ab,\xb} d\xb dy\right] &&\text{(by Lemma~\ref{lemma:pseudo-prob})}\\
& = \EE_{\AB}~\left[\int\int   \log f_\theta(y|\ab) w(\ab,\xb) \prob{y,\ab,\xb} d\xb dy\right] \\
& = \EE_{(y,\ab,\xb) \sim f}w(\ab,\xb) \log f_\theta(y|\ab)&&\text{($\AB$ is uniform)}\\
& \approx \frac{1}{N} \sum_{(y,\ab,\xb) \in \mathcal{D}} w_\phi(\ab,\xb) \log f_\theta(y|\ab), 
\end{align*}
where $N$ represents the sample size, and  $w_\phi(\ab,\xb)$ denotes the learned subject-specific IPTW, parameterized by $\phi \in \Phi$, which takes the form:
    \begin{equation}
        w_\phi(\ab,\xb) = \frac{1}{\prod_{\tau=t-d+1}^{t} f_\phi(a_{\tau}|\ab_{\tau-1},\xb_{\tau})}.
    \end{equation}

\label{append:proof-prop-1}
Note that we choose $\AB$ to be uniformly distributed for a simplified representation. In practice, $\AB$ can also be taken from the observed distribution (and hence may be imbalanced across treatment combinations). This will not affect the optimal parameters as long as the form of the generator is flexible enough, which is a common assumption.

\section{Connection to related methods } \label{append:connection-other}

\paragraph{Plug-in density estimation} Plug-in approaches have been commonly used to estimate the counterfactual density in the static setting\citep{bickel2001inference, kim2018causal, Kennedy23} and can be extended to our time-varying setting via direct application of Lemma~\ref{lemma:pseudo-prob}. However, this practice could be problematic when the sample size is large as it requires averaging the entire observed dataset for each evaluation of $y$. Instead, we circumvent this computational challenge by approximating the counterfactual density using a proxy conditional distribution $f_\theta(\cdot|\ab)$ which is represented by a generative model,  $ g_\theta(z, \ab)$. 

\paragraph{(Semi)parametric density estimation.}
Our framework uses a conditional generator, $ g_\theta(z, \ab)$, to approximate the proxy conditional distribution $f_\theta(\cdot|\ab)$ under all $\ab$. This differs from the parametric/semi-parametric density estimation approaches in \cite{Kennedy23} and \cite{melnychuk2023normalizing}, which directly estimates $f_\theta(\cdot|\ab)$ for each $\ab$. The major advantage of our generative framework is its applicability in high-dimensional outcomes: it is computationally prohibitive to estimate high-dimensional densities. Therefore, we generate samples that represent this high-dimensional counterfactual distribution, instead of estimating it. This is a common practice in generative models, such as GAN \citep{goodfellow2014generative} and VAE \citep{kingma2013auto}, where the generator learns to generate diverse image samples without estimating the density underlying the image distributions. Another advantage of our approach is that it trains only a single model for all treatment combinations, and has the potential to generalize to continuous treatments. Furthermore, our Proposition \ref{prop:pesudo-objective} is more sample efficient as compared to the parametric/semi-parametric density estimation approaches. The framework in \cite{melnychuk2023normalizing}, when extended to the time-varying scenario using IPTW, requires integrating the log-likelihood of the density model over both the observed samples and the outcome space $\mathcal{Y}$, see (\ref{eq:melnychuk}). In practice, this will require performing a Monte Carlo sampling of $Y$ for each gradient step to optimize (\ref{eq:melnychuk}), which can be prohibitive when $\mathcal{Y}$ is high-dimensional. Our proposed loss function in Proposition \ref{prop:pesudo-objective}, on the other hand, only requires computing the weighted log-likelihood over observed samples which is easy to implement. Therefore, our Proposition \ref{prop:pesudo-objective} can be viewed as a novel reformulation of (\ref{eq:melnychuk}) into a generative training framework that enhances the scalability of model training for high-dimensional outcomes.

    \begin{equation}
       \EE_{y \sim f_{\ab}} \log f_{\theta}(y)  \approx \int_{y\in\mathcal{Y}}\log f_{\theta}(y) \frac{1}{N}\sum_{(y,\ab,\xb) \in \mathcal{D}}  w_\phi(\ab,\xb) f(y, \ab, \xb)dy.
        \label{eq:melnychuk}
    \end{equation}

\paragraph{Doubly-robust density estimators.}

Doubly-robust density estimators have proven successful in directly estimating the counterfactual density in the static setting \citep{Kennedy23,melnychuk2023normalizing}. In theory, one may extend our IPTW-based framework to a doubly-robust framework to ensure robustness.  We opted for an IPTW only based framework because of practical challenges in combining IPTW with a outcome-based model, such as G-computation \citep{robins2008estimation,li2021g}, for high-dimensional outcomes.  When $Y$ is potentially high-dimensional, correct estimation of the outcome model using methods such as G-net might be challenging. This is due to the need to estimate both the conditional outcome distribution $f(y_t|\xb_t,\ab_t)$ and the conditional covariate distribution $f(x_t|\xb_{t-1},\ab_{t-1})$, where the first term involves a high-d outcome and the second term involves estimating a continuous density. As a result, we have also shown that the outcome-based method, \texttt{G-Net}, underperforms than our IPTW based framework. Therefore, we opted for the IPTW-based approach in proposition \ref{prop:pesudo-objective}. We remain open about the possibility of adopting the doubly-robust approach for future work, especially if insights emerge regarding the development of an outcome-based method for accurately approximating high-dimensional, time-varying counterfactual distributions.


\section{Derivation and implementation of the classifier-free guided diffusion}
\label{app:conditional-ddpm}
Diffusion models gradually add noise to the original input (the outcome, $y\coloneqq y_0$) through $S$ forward diffusion steps and learn to denoise a random noise in the reverse diffusion steps. The forward process is defined as:
\[
    q(y_{s}|y_{s-1}) \coloneqq \mathcal{N}(\sqrt{1 - \gamma_{s}}y_{s-1}, \gamma_{s}I), \quad q(y_{1:S}|y_0) \coloneqq  \prod_{s=1}^{S} q(y_{s}|y_{s-1}),
\]
where $y_s$ represents the noise version of the outcome at the $s$-th step. The reverse process is defined as:
\[
    p_{\theta}(y_{0:S}) \coloneqq p(y_{S}) \prod_{s=1}^{S} p_{\theta}(y_{s-1}|y_{s}), 
\]
\[
     p_{\theta}(y_{s-1}|y_{s}) \coloneqq \mathcal{N}(\mu_{\theta}(y_{s}, s), \Sigma_{\theta}(y_{s}, s)),
\]
where the mean $\mu_{\theta}$ is represented by a neural network and the covariance $\Sigma_{\theta}$ is set to be $\gamma_s I$. 
The mean function $\mu_{\theta}$ can be trained by maximizing the variational bound on the log-likelihood of the data $y$:
\[
    \mathbb{E}[\log p_{\theta}(y)] \geq \mathbb{E}_q\left[ \log p_{\theta}(y_S) + \sum_{s\geq1}\log\frac{p_{\theta}(y_{s-1}|y_s)}{q(y_s|y_{s-1})} \right].
\]
To optimize the above objective, we note the property of the forward process that allows the closed-form distribution of $y_s$ at an arbitrary step $s$ given the data $y$: denoting $\lambda_s \coloneqq 1-\gamma_s$ and $\bar{\lambda}_s \coloneqq \prod_{r=1}^s\lambda_s$, we have $q(y_s|y) = \mathcal{N}(\sqrt{\bar{\lambda}_s}y, (1-\bar{\lambda}_s)I)$.
The variational bound can be therefore expressed by the KL divergence between a series of Gaussian distributions of $y_{1:S}$ and $y$ \citep{sohl2015deep}, leading to the following equivalent training objective:
\[
    \mathbb{E}_q\left[ \sum_{s>1}\frac{1}{2\gamma_k}\|\tilde{\mu}_s(y_s, y) - \mu_{\theta}(y_s, s)\|^2 \right] + C,
\]
where $\tilde{\mu}_s(y_s, y) = \frac{\sqrt{\bar{\lambda}_{s-1}}\gamma_s}{1 - \bar{\lambda}_s}y + \frac{\sqrt{\lambda_s}(1 - \bar{\gamma}_{s-1})}{1 - \bar{\gamma}_s}y_s$ and $C$ is a constant that does not depend on $\theta$. Re-parametrizing $y_s = \sqrt{\bar{\lambda}_s}y + \sqrt{1-\bar{\lambda}_s}\epsilon$ with $\epsilon\sim \mathcal{N}(0, I)$, the model is trained on the final variant of the variational bound:
\[
   \mathbb{E}_{s ,y,\epsilon_s}||
  \epsilon_s - \epsilon_\theta(\sqrt{\bar{\lambda}_s}y+\sqrt{1-\bar{\lambda}_s}\epsilon_s,s)||^2.
\]
Conditioning on the treatment $\ab$, the objective becomes:

\[
   \mathbb{E}_{s \sim [1,S],y\sim f(y|\ab),\epsilon_s}||
  \epsilon_s - \bar{\epsilon}_\theta(\sqrt{\bar{\lambda}_s}y+\sqrt{1-\bar{\lambda}_s}\epsilon_s,s,\ab)||^2,
\]
where the score model  $\bar{\epsilon}_\theta(y_s,s,\ab) = (w+1) \epsilon_\theta(y_s,s,\ab)-w\epsilon_\theta(y_s,s)$ is a linear combination of the conditional and unconditional score functions for the classifier-free guidance \citep{ho2022classifier}.

\section{Derivation and implementation details of variational learning}
\label{append:variational-learning}

\paragraph{Derivation of the proxy conditional distribution}
Now we present the derivation of the log conditional probability density function (PDF) in (\ref{eq:variational-lower-bound}).
To begin with, it can be written as:
\[
    \log f_{\theta}(y|\ab) = \log \int p_{\theta}(y, z|\ab)dz,
\]
where $z$ is a latent random variable.
This integral has no closed form and can be usually estimated by Monte Carlo integration with importance sampling, \ie, 
\[
    \int p_{\theta}(y, z|\ab)dz = \EE_{z\sim q(\cdot |y, \ab)}\left[\frac{p_{\theta}(y, z|\ab)}{q(z|y, \ab)}\right].
\]
Here $q(z|y, \ab)$ is the proposed variational distribution, where we can draw sample $z$ from this distribution given $y$ and $\ab$. 
Therefore, by Jensen's inequality, we can find the evidence lower bound (ELBO) of the conditional PDF:
\[
    \log f_{\theta}(y|\ab) = \log \EE_{z\sim q(\cdot|y, \ab)}\left[\frac{p_{\theta}(y, z|\ab)}{q(z|y, \ab)}\right] \geq \EE_{z\sim q(\cdot|y, \ab)} \left[ \log \frac{p_{\theta}(y, z|\ab)}{q(z|y, \ab)}\right].
\]
Using Bayes rule, the ELBO can be equivalently expressed as:
\begin{align*}
& \EE_{z\sim q(\cdot|y, \ab)} \left[ \log \frac{p_{\theta}(y, z|\ab)}{q(z|y, \ab)}\right] \\
=& \EE_{z\sim q(\cdot|y, \ab)} \left[ \log \frac{p_{\theta}(y|z, \ab)p_{\theta}(z|\ab)}{q(z|y, \ab)}\right] \\ 
=& \EE_{z\sim q(\cdot|y, \ab)} \left[ \log \frac{p_{\theta}(z|\ab)}{q(z|y, \ab)}\right] + \EE_{z\sim q(\cdot|y, \ab)} \left[ \log p_{\theta}(y|z, \ab) \right] \\
=& - D_\text{KL}(q(z|y, \ab) || p_{\theta}(z|\ab)) + \EE_{z\sim q(\cdot|y, \ab)} \left[ \log p_{\theta}(y|z, \ab) \right].
\end{align*}


\paragraph{Implementation details} 

For the KL-divergence term in the ELBO (\ref{eq:variational-lower-bound}), both $q(z|y, \ab)$ and $p_{\theta}(z|\ab)$ are often modeled as Gaussian distributions, which allows us to compute the KL divergence of Gaussians with a closed-form expression. In practice, we introduce two additional generators, including the encoder net $g_\text{encode}(\epsilon, y, \ab)$ and the prior net $g_\text{prior}(\epsilon, \ab)$, respectively, to represent $q(z|y, \ab)$ and $p_\theta(z|\ab)$ as transformations of another random variable $\epsilon \sim \mathcal{N}(0, I)$ using reparameterization trick \citep{sohl-dickstein2015deep}. A common choice is a simple factorized Gaussian encoder. For example, the approximate posterior $q(z|y, \ab)$ can be represented as:
\[
    q(z|y, \ab) = \mathcal{N}(z; \mu, \text{diag}(\Sigma)),
\]
or 
\[
    q(z|y, \ab) = \prod_{j=1}^r q(z_j|y, \ab) = \prod_{j=1}^r \mathcal{N}(z_j; \mu_j, \sigma_j^2).
\]
The Gaussian parameters $\mu = (\mu_j)_{j=1,\dots,r}$ and $\text{diag}(\Sigma) = (\sigma_j^2)_{j=1,\dots,r}$ can be obtained using reparameterization trick via an encoder network $\phi$:
\begin{align*}
    (\mu, \log \text{diag}(\Sigma)) =&~ \phi(y, \ab),\\
    z =&~ \mu + \sigma \odot \epsilon, 
\end{align*}
where $\epsilon \sim \mathcal{N}(0, I)$ is another random variable and $\odot$ is the element-wise product.  Because both $q(z|y_i, \ab_{i-1})$ and $p_{\theta}(z|\ab_{i-1})$ are modeled as Gaussian distributions, the KL divergence can be computed using a closed-form expression. 

The log-likelihood of the second term can be implemented as the reconstruction loss and calculated using generated samples. Maximizing the negative log-likelihood $p_\theta(y|z, \ab)$ is equivalent to minimizing the cross entropy between the distribution of an observed outcome $y$ and the reconstructed outcome $\widetilde{y}$ generated by the generative model $g$ given $z$ and the history $\ab$.  


\section{Additional experiment details}

\subsection{Baselines}\label{append:baselines}
Here we present an additional review of each baseline method in the paper as well as implementation details. 

\paragraph{Marginal structural model with a fully-connected neural network (\texttt{MSM+NN})} We include the classic \texttt{MSM+NN} proposed in \citep{robins1994estimation,robins1986new}.
This classical framework assumes that the counterfactual mean of the outcome variable can be represented as a linear function of the treatments. 
We use this model while replacing the linear model with a $3$-layer fully-connected neural network, $g_\text{msm}$. This serves as a deterministic baseline for our generative framework.
We learn the \texttt{MSM+NN} using stochastic gradient descent with a weighted loss function:
\[
    \sum_{(y,\ab,\xb) \in \mathcal{D}}w_\phi(\ab,\xb) (y-g_\text{msm}(\ab))^2.
\]
To establish a fair comparison, we train the \texttt{MSM+NN} using an identical training size to that of the \texttt{MSCVAE} model. We train the \texttt{MSM+NN} for $1,000$ epochs with a learning rate of $0.01$. However, it's important to note that in this particular setup, our capacity is limited to estimating the mean instead of the entire distribution. For computing the Wasserstein distance in the full-synthetic experiments, we treat the \texttt{MSCVAE} samples as coming from a degenerate distribution at its predicted value.

\paragraph{Conditional variational autoencoder (\texttt{CVAE})}
We include a vanilla conditional variational autoencoder (\texttt{CVAE}) with an architecture identical to that of the \texttt{MSCVAE}, but excluding IPTW weighting.  
The \texttt{CVAE} is a widely-used type of conditional generative model that has found applications in various tasks, including image generation \citep{mishra2018generative,sohn2015learning}, neural machine translation \citep{pagnoni2018conditional}, and molecular design \citep{lim2018molecular}. 
To train the \texttt{CVAE}, we follow the same procedure as \texttt{MSCVAE}, with the exception that we replace the loss function with the unweighted version of (\ref{eq:weighted-log-likelihood}):
\[
    \sum_{(y,\ab,\xb) \in \mathcal{D}} \log f_\theta(y|\ab),
\]
where $f_\theta(\cdot)$ is the conditional distribution represented by the \texttt{CVAE}.

\paragraph{Kernel density estimator (\texttt{KDE})}
We use a Gaussian kernel density estimator \citep{rosenblatt1956remarks}  to estimate the empirical conditional distribution from the observed data. This is achieved by running KDE on the observed outcomes with the same treatments, \ie,
\[
f_{\ab}\approx g_\text{kde}(y|\AB=\ab),
\]
where $ g_\text{kde}(\cdot)$ is the KDE estimator. We learn the KDE with bandwidth set to $0.5$, $1$, $1.5$, and $2$, respectively, and report the metrics with bandwidth $=0.5$ as the optimal results.

\paragraph{Semi-parametric Plug-in method based on pseudo-population (\texttt{Plugin+KDE}) }  We include a baseline using Lemma~\ref{lemma:pseudo-prob} as a plugin estimator by following the semi-parametric KDE approach in \cite{melnychuk2023normalizing}. Specifically, we rewrite Lemma~\ref{lemma:pseudo-prob}  as:
\[
\cprob{\ab}{y} \approx \sum_{(y,\ab,\xb) \in \mathcal{D}} \mathbbm{1}\{\AB = \ab\} w_\phi(\ab,\xb) \prob{y, \AB, \xb}.
\]
To estimate the right-hand side of the equation, we performed KDE on $y|\AB=\ab$ where each sample tuple $(y,\ab,\xb)$ is weighted by its IPTW, $w_\phi(\ab,\xb)$, for each $\AB =\ab$ separately. The bandwidth is set to be the same as in \texttt{KDE}.

\paragraph{G-Net (\texttt{G-Net})}
We implement \texttt{G-net} proposed in \citep{li2021g} based on G-computation. For our experiment setting, at each time step $t \in [T]$, we designed the conditional covariates block, the history representation block, and the final conditional outcome block as a 3-layer fully connected neural network respectively. The types of blocks are interconnected to form sequential net structures across different time steps, followed by a conditional outcome block at the end, which has a 2-layer structure. This makes the \texttt{G-net} model include a total of $(2 \times d) + 1$ blocks. The loss function is the sum of the mean squared error:
\[
\sum_{(\xb, y) \in \mathcal{D}} (\widehat{\xb} - \xb)^2 +  (\widehat{y} - y)^2,
\]
where $\widehat{\xb}$ and $\xb$ are the predicted and groundtruth covariate history, while $\widehat{y}$ and $y$ are the predicted and groundtruth outcome.
Following the original literature, we impose a Gaussian parametric assumption over the underlying counterfactual distribution, and introduce prediction variability by adding Gaussian noise whose variance is empirically estimated from the residuals between the predicted and groundtruth outcomes.

\paragraph{Diffusion (\texttt{Diffusion})}

We include a guided diffusion model (\texttt{Diffusion}) \citep{ho2022classifier} with an architecture identical to that of the \texttt{MSDiffusion}, but excluding IPTW weighting.  
The \texttt{Diffusion} is a commonly-used type of conditional generative model with tremendous success in generating high-quality samples such as images \citep{ho2020denoising,  ho2022classifier,rombach2022high,yang2023diffusion}
\citep{ho2022classifier}.
To train the \texttt{Diffusion}, we follow the same procedure as \texttt{MSDiffusion}, with the exception that we replace the loss function with the unweighted version of (\ref{eq:weighted-log-likelihood}):
\[
    \sum_{(y,\ab,\xb) \in \mathcal{D}} \log f_\theta(y|\ab),
\]
where $f_\theta(\cdot)$ is the conditional distribution represented by \texttt{Diffusion}.

\paragraph{Counterfactual Recurrent Network (\texttt{CRN})} We include \texttt{CRN} proposed in \cite{bica2019estimating} as our baseline for the fully synthetic experiment. The model consists of a recurrent encoder and decoder structure consisting of LSTM blocks. The treatment classifier $G_a$ and the outcome predictor network $G_y$ are constructed as fully-connected linear layers with softmax and linear output layers respectively. $G_a$ additionally contains a gradient reversal layer \cite{ganin2016domain}. The balanced representation is computed from a linear layer with ELU activation function. Two modifications were made: (1) We use standard dropout to the LSTM instead of variational dropout, which will not negatively affect modelling power, and (2) We limit the decoder training procedure to use a subset of mini-batches of splited data, which is set as a hyperparameter and can be tuned to balance computational efficiency and model performance.

\begin{table*}[t!]
 \caption {Coefficients of the linear model in synthetic data generation}
 \centering
 \begin{adjustbox}{max width=1\textwidth}
 \begin{tabular}{ c   c   c   c } 
\toprule[1pt]\midrule[0.3pt]
 \textbf{} & {$\alpha$}& {$\beta$} & {$\gamma$}\\
\hline
$d=1$ & $(-3,2,0,-1,0)$ & $(-0.5,0.5,-0.5,0.5)$ & $(0,1,-1)$\\

$d=3$ & $(-1, 3,6,12,0, 0.5,1,2,0)$ & $(-0.5,0.5,-0.5,0.5,-0.5,0.5,-0.5,0.5)$ &  $(-1,0.5,1,1.5,-0.5,-1,-1.5)$\\

$d=5$ & $(-1, 0.5,1,3,6,12,0, 0.05,0.1, 0.5,1,2,0)$ & $(-0.5, 0.5,-0.5,0.5,-0.5,0.5, -0.5,0.5,-0.5,0.5,-0.5,0.5)$  &  $(-1,0.05,0.1, 0.5,1,1.5,-0.05,-0.1, -0.5,-1,-1.5)$\\

\midrule[0.3pt]\bottomrule[1pt]
\end{tabular}
 \end{adjustbox}
 \label{append:tab-coeff}
\end{table*}

\subsection{Experiment set-up}
\label{append:set-up}

\paragraph{Neural network architecture for \texttt{MSCVAE}}

The counterfactual generator $g_\theta$, the IPTW $w_\phi$, and the encoder network $g_\text{encode}$ share the same two-layer fully-connected network architecture with ReLU activation. The layer width is set to $1,000$, and the length of the latent variable $z$  is set to $r$ which is determined by the specific synthetic experiment setting: $r=5$ for $d=1$ and $d=3$, $r=10$ for $d=5$ and all the semi-synthetic and real data. For $g_\text{encode}$, the fully-connected networks map the $d+1$ dimensional input vector (consisting of a $d$-dimensional treatment and $1$-dimensional response) to the $r$-dimensional latent representation. For $g_\theta$, the fully-connected networks map the $r+d$ dimensional input vector (consisting of a $d$-dimensional treatment and $r$-random noise) to the $1$-dimensional generated counterfactual outcome. For $w_\phi$, the fully-connected networks map the $2d$-dimensional input vector (consisting of a $d$-dimensional treatment and $d$-dimensional covariate) to the $1$-dimensional conditional probability. We use a Sigmoid output layer for $w_\phi$ to ensure the output falls within $[0,1]$. We set the batch size to $256$ and the number of training epochs to $200$ for training all the models in both synthetic and real data settings. The learning rate was set to $10^{-3}$ with a linear step-wise learning rate scheduler (Pytorch learning rate scheduler function \texttt{StepLR}) to ensure stable convergence of the learning process.

\paragraph{Neural network architecture for \texttt{MSDiffusion}}
The IPTW $w_\phi$ share the same two-layer fully-connected network architecture as in \texttt{MSCVAE} with ReLU activation. The layer width is set to $1,000$. For the score model  $\bar{\epsilon}_\theta(y_s,s,\ab)$, we used separate architectures for the semi-synthetic TV-MNIST and COVID-19 datasets. For TV-MNIST, we used a U-net\citep{ronneberger2015u}  with two convolution layers (feature dimensionality is $256$ and $512$), two up-convolution layers (feature dimensionality is $512$ and $256$), and GELU activation \citep{hendrycks2016gaussian}. For the COVID-19 datasets, we used an autoencoder with two layers in both the encoder and decoder (both layer width and length of the latent variable set to $512$), and GELU activation \citep{hendrycks2016gaussian}. For both score models, we used a two-layer MLP to embed the treatment $\ab$ and the time step $s$ into a $256$-dimensional space. We set the batch size to $256$ and the number of training epochs to $50$ for training. The learning rate was set to $10^{-4}$ with a linear step-wise learning rate scheduler (Pytorch learning rate scheduler function \texttt{StepLR}) to ensure stable convergence of the learning process. During the generation stage, we set the guidance strength $w=3$ for the TV-MNIST dataset and $w=2$ for the COVID-19 datasets.

\subsection{Experiment metrics}
\label{append:metrics}
To quantify the quality of the approximated counterfactual distributions, we used the following metrics:

\paragraph{Mean} This is the difference between the empirical mean of the evaluated samples.

\paragraph{1-Wasserstein Distance} We used the earth mover's distance, which is defined as:

\[
l_1(u,v) = \inf \limits_{\pi \in \Gamma (u,v)} \int_{\Omega\times \Omega }|x-y|d\pi(x,y),
\]
where $\Gamma (u,v)$ is the joint probability distributions for the groundtruth and learned counterfactual distributions, and $\Omega$ is the space of each distribution.

\paragraph{FID*} 
Both semi-synthetic datasets have high-dimensional outcomes, making comparisons using the mean or Wasserstein distance of the distributions less interpretable. A common approach in the generative model community is FID (Fréchet inception distance ). In summary, FID uses a pre-trained neural network (frequently the inception v3 model) to obtain a feature vector for each sample, generated for groundtruth. The feature vector is the activation of the last pooling layer prior to the output layer of the pre-trained network. The feature vectors are then summarized as multivariate Gaussians by computing their mean and covariances. The distance between the generated or groundtruth image distribution is then computed by calculating the 2-Wasserstein distance between two sets of Gaussians. A lower FID score represents a more realistic distribution for the generated images.


Since FID is not specifically designed for our TV-MNIST and semi-synthetic COVID-19 datasets, we propose to use FID* by following a similar idea of FID. For the semi-synthetic COVID-19 dataset, we first use a geographical projector to map each sample in $\mathbb{R}^{67}$ to a $2$-d coordinate of latitude and longitude of the maximal entry. The projection serves as the purpose of the pre-trained network in the original FID because it outputs the `hotspot' of the generated counterfactual outcome vector. For each treatment combination, we then computed the 2-Wasserstein distance of the projection between the hotspot coordinates of the generated and counterfactual samples. A lower FID* score represents the generated samples have a similar spatial distribution compared to the counterfactual ones.

For the TV-MNIST dataset, we use a $3$-layer fully-connected neural network pre-trained to classify MNIST images. This network serves as the purpose of the pre-trained network in the original FID because it represents the semantic information (the digit label) of the $784$-dimensional outcome variables. For each treatment combination, we then project the $784$-dimensional samples into a $1$-dimensional label space using the pre-trained MNIST classifier and compute the 1-Wasserstein distance of the projection between the generated and counterfactual samples. A lower FID* score represents the generated samples have a similar semantic distribution (in terms of the digit labels) compared to the counterfactual ones.

\subsection{Fully Synthetic data}\label{append:synthetic}

\begin{algorithm}[t!]
\begin{algorithmic}
    \STATE {\bfseries Input:} 
    Generated trajectory of a single subject: $\{(y_t,x_t,a_t)\}_{t=1,\cdots,T}$. 
    \STATE {\bfseries Initialization:} Given the treatment history $\ab_T = \ab$.
    \FOR{$\tau=T-d+1:T$}
    \STATE 1. Generate the covariate $x_{\tau}$ based on $\ab_{\tau-1}$ and $\xb_{\tau-1}$ according to (\ref{append:synthetic-X}).
    \STATE 2. Update the covariate $\xb_{\tau} \gets x_{\tau}$.
    \ENDFOR
    \STATE Generate $Y(\ab)$ based on  $\ab_T$ and $\xb_{T}$ according to (\ref{append:synthetic-Y}).
     
    \STATE {\bfseries return} $Y(\ab)$ \;
    
\end{algorithmic}
\caption{Algorithm for obtaining a counterfactual sample }
\label{append:algo:counterfactual}
\end{algorithm}

In this section, we provide an overview of the procedures for generating synthetic data.
Our goal is to evaluate the performance of the proposed \texttt{MSCVAE} method and compare it to baseline approaches in the context of time-varying treatments. 
We follow the classic setting in \citep{robins1999estimation} and simulate time series data with time-varying treatments and covariates. 
The presence of the time-varying confounders serves as an appropriate testbed for comparing MSM-based models to the baselines. 
To be specific, we generate three synthetic datasets with varying levels of historical dependence denoted as $d$. Each dataset consists of 10,000 trajectories, which represent recorded observations of individual subjects. These trajectories comprise 100 data tuples, encompassing treatment, covariate, and outcome values at specific time points.
The causal relationships between these variables are visually depicted in Figure~\ref{fig:graphical-model}. For each time trajectory of length $T$, the datasets are generated based on the following equations:
\begin{align}
X_0 &~\sim \text{uniform}(0,1),\\
X_t&~=\gamma_0 +  \sum_{\tau=t-d}^{t-1}\gamma_{t-\tau}A_{\tau} +  \sum_{\tau=t-d}^{t-1}\gamma_{d+t-\tau}X_{\tau} \label{append:synthetic-X},\\
\mathbb{P}\{A_t=1\} &~= \sigma(\beta_0 +  \sum_{\tau=t-d}^{t-1}\beta_{t-\tau}A_{\tau} +  \sum_{\tau=t-d}^{t}\beta_{d+t-\tau+1}X_{\tau}) \label{append:synthetic-A},\\ 
Y_t &~=\alpha_0 + \sum_{\tau=t-d}^{t}\alpha_{t-\tau+1} A_{\tau} + \sum_{\tau=t-d}^{t}\alpha_{d+t-\tau+2}X_{\tau} +\epsilon,
\label{append:synthetic-Y}
\end{align}
where $\epsilon \sim \mathcal{N}(0,0.05)$ is the observation noise and $\sigma(\cdot)$ is a Sigmoid function. Note that here we count from $t-d$ instead of $t-d+1$ to align $d$ with the notation $K$ in \cite{robins2008estimation}. The specific coefficients are set according to the values in Table~\ref{append:tab-coeff} to ensure the generation of valid synthetic data distributions with diversity: adjusting $\beta_0$ will change the balance of the treatment combinations:  when keeping the remaining $\beta$ coefficients, treatment variables $\ab$, and covariates $\xb$ unchanged, a smaller value of $\beta_0$ reduces the probability of treatment exposure, \ie, $\mathbb{P}(A_t=1)$. Consequently, this lower probability of treatment exposure results in a decrease in the occurrence of treatment combinations with exposures, leading to an imbalanced ratio among different treatment combinations. In Figure.~\ref{fig:synthetic-exps}, we set $\beta_0=-0.5$ which results in an approximated balanced number of samples per treatment combination.  In Appendix~\ref{append:additional}, we include a figure by setting $\beta_0=-2$, as a visualization of imbalanced treatment combinations.



To ensure the validity of our synthetic data generation process, we verify that the three assumptions in the Methodology section are satisfied. Assumptions 1 and 3 are naturally met because the ground truth model guarantees that the counterfactual outcome equals the observed outcome and that there are no unmeasured confounders. As for assumption 2, since the conditional probability of treatment is the Sigmoid function applied to a finite linear combination of historical treatments and covariates, it will always be positive.


Once the synthetic data is generated, we obtain counterfactual distributions to assess the performance of our proposed method. Specifically, we use the synthetic data to obtain samples from the counterfactual outcome distribution, ${Y(\ab)}$, for any given treatment combination $\ab$. This is achieved by iteratively fixing the treatment sequence in the time series and generating the covariates and response variables according to equations (\ref{append:synthetic-X}) and (\ref{append:synthetic-Y}) for each of the $10,000$ trajectories. The detailed procedure for obtaining a single counterfactual outcome sample is summarized in Algorithm \ref{append:algo:counterfactual}.

\subsection{Semi-synthetic time-varying MNIST data}\label{append:tv-mnist}

We provide a benchmark based on the MNIST dataset. Specifically, the outcomes are MNIST images ($m=784$). First, we compute a one-dimensional summary, the $\phi$ score \citep{jesson2021quantifying}, using each MNIST image. The $\phi$ value of an image depends on its average light intensity and its digit label. We refer the readers to \cite{jesson2021quantifying} for the details on computing $\phi$. Here we set the length of history dependence, $d$, to $3$. We then define a linear model of $1$-dimensional latent process to \ref{append:synthetic} and simulate $1,000$ trajectories of the $(X,A,Y)$ tuples of $100$ time points according to the following equations: 

\begin{align}
X_0 &~\sim \text{uniform}(0,1),\\
X_t&~=\gamma_0 +  \sum_{\tau=t-2}^{t-1}\gamma_{t-\tau}A_{\tau} +  \sum_{\tau=t-2}^{t-1}\gamma_{t-\tau+2}X_{\tau} ,\\
\mathbb{P}\{A_t=1\} &~= \sigma(\beta_0 +  \sum_{\tau=t-2}^{t-1}\beta_{t-\tau}A_{\tau} +  \sum_{\tau=t-2}^{t}\beta_{t-\tau+3}X_{\tau}),\\ 
\phi_t &~=0.5 \left \lceil{10\sigma(\alpha_0 + \sum_{\tau=t-2}^{t}\alpha_{t-\tau+1} A_{\tau} + \sum_{\tau=t-2}^{t}\alpha_{t-\tau+4}X_{\tau})-0.6}\right \rceil, \\
Y_t  &~ \sim \{\texttt{MNIST}(i): i= \argmin |\phi_i-\phi_t|   \}, 
\end{align}
where $\sigma(\cdot)$ is a Sigmoid function, $\left \lceil{\cdot}\right \rceil$ is the ceiling function, and $\texttt{MNIST}(i)$ represents the MNIST image indexed by $i$. The coefficients are set according to Table~\ref{append:tab-coeff} to ensure the generation of diverse data distributions. We generate the counterfactual samples according to Algorithm~\ref{append:algo:counterfactual} by replacing the corresponding propensity and outcome models with the formulations above. The generated observations and counterfactual samples under the same treatment combinations may correspond to MNIST images of different labels. This way we can qualitatively assess the performance of an algorithm by comparing the labels of the MNIST images it generates against the counterfactual samples, as in as in Figure.~\ref{fig:tvmnist}.

\subsection{COVID-19 data}\label{append:real-data}

\begin{table*}[t]
 \caption {Real data description}
 \centering
 \begin{adjustbox}{max width=1\textwidth}
 \begin{tabular}{ l   l   l  l l l l    } 
\toprule[1pt]\midrule[0.3pt]
 \textbf{Name} & {\textbf{Description}}& Min & Max & Mean  & Median  & Std\\
\hline
$Y$ & county-wise incremental new cases count ( $\log_{10}$)& $0$& $1.15\times 10^{-1}$& $2\times 10^{-3}$& $1\times 10^{-3}$& $2.7\times 10^{-3}$\\

$A$ & county-wise mask mandate &  $0$ &  $1\times 10^{0}$ &  $5.35\times 10^{-1}$ &  $1\times 10^{0}$ &  $4.99\times 10^{-1}$\\

$X^{(0)}$ &county-wise incremental death cases count ( $\log_{10}$)&  $0$& $3.12\times 10^{-3}$& $3\times 10^{-4}$& $0$& $9\times 10^{-5}$\\
$X^{(1)}$ & county-wise average retail and recreation &  $-5.45\times 10^1$& $2.23 \times 10^1$& $-4.27\times 10^{0}$& $-3.33\times 10^{0}$& $6.16\times 10^{0}$\\

$X^{(2)}$ & county-wise symptom value &  $0$& $3.23\times 10^1$& $9.3 \times 10^{-1} $& $8.1 \times 10^{-1}$& $5.1 \times 10^{-1}$\\
\midrule[0.3pt]\bottomrule[1pt]
\end{tabular}
 \end{adjustbox}
 \label{append:tab-real}
\end{table*}

Since both the semi-synthetic Pennsylvania COVID-19 mask data and the real nationwide COVID-19 mask datasets are based on the same set of aggregated sources. We first introduce the data sources and then include the details of each dataset respectively.

The real data used in this study comprises COVID-19-related demographic statistics collected from $3,219$ counties across $56$ states/affiliated regions of the United States. The data covers a time period from $2020$ to $2022$. We obtained the data from reputable sources including the U.S. Census Bureau \citep{uscensus}, the Center for Disease Control and Prevention \citep{centers2021us}, Google \citep{google}, the CMU DELPHI group's Facebook survey \citep{delphi}, and the New York Times \citep{nyt}.
 To capture a relevant time window for analysis, we set the history dependence length $d$ to $3$, as most COVID-19 symptoms tend to subside within this timeframe \citep{maltezou2021post}. 

In our analysis, the treatment variable $A$ is the state-wise mask mandate indicator variable. A value of $0$ indicates no mask mandate, while a value of $1$ indicates the enforcement of a mask mandate. Notably, we observe a pattern in the data where mask mandates are typically implemented simultaneously across all counties within a state. This synchronization justifies the use of the state-wise mask mandate count as the treatment variable. As for the covariates $X$, we choose the county-wise incremental death count, state-wise the average retail and recreation metric (representing changes in mobility levels compared to a baseline, which can be negative), the state-wise symptom value, and the state-wise vaccine dosage.

\paragraph{Pennsylvania COVID-19 mask mandate data}

For the semi-synthetic dataset, we specifically look at the data within the state of Pennsylvania because of its long records spanning $106$ weeks from $2020$ to $2021$. We set the four state-level covariates (per $100$K people): the number of deaths, the average retail and recreation mobility, the surveyed COVID-19 symptoms, and the number of administered COVID-19 vaccine doses. We set the county-level incremental death count to the state level by computing a state average. We set the state-level mask mandate policy as the treatment variable, and the county-level number of new COVID-19 cases (per $100$K) as the outcome variable, resulting in $m=67$ since there are $67$ counties in the state of Pennsylvania. We simulate $2,000$ trajectories of the $(X,A,Y)$ tuples of $300$ time points (each point corresponding to a week) according to the following formula:
\begin{align}
X_0 &~\sim \text{Real-World}(\cdot),\\
X_t&~=\hat{\mathbb{P}}(X_t|\AB_t,\XB_t),\\
\mathbb{P}\{A_t=1\} &~= \sigma(\beta_0 +  \sum_{\tau=t-2}^{t-1}\beta_{t-\tau}A_{\tau} +  \sum_{\tau=t-2}^{t}\beta_{t-\tau+3}X_{\tau}) ,\\ 
Y_t^{base}&~= -0.2A_{t-2} -0.15A_{t-1} -0.1A_{t}+0.45+\epsilon, \\
\mathbb{P}(L_t=1) &~= \text{Bernoulli} (\prod_{j=1}^{4}X_{t}(j)),\\
Y_t(s)&~=Y_t^{base}+\begin{cases}
 \log(\mathcal{N}(s,\mathbf{\mu}=[40.009, -75.133]^T, \Sigma=\mathbf{I}));  \; \\\text{if } L_t=1,\\      
 \log(\mathcal{N}(s,\mathbf{\mu}=[40.470, -79.980]^T, \Sigma=\mathbf{I})); \; \\\text{otherwise}.
\end{cases}
\end{align}
where $\hat{\mathbb{P}}(\cdot)$ is learned with a $2$-layer fully-connected neural network using the real data, $\epsilon \sim \mathcal{N}(0,0.001)$ is the observation noise, $s$ is the $2$-dimensional coordinate of a entry (county) in $Y_t$, $\sigma(\cdot)$ is a Sigmoid function. All other coefficients are set according to Table~\ref{append:tab-coeff} to ensure the generation of diverse data distributions. We generate the counterfactual samples according to Algorithm~\ref{append:algo:counterfactual} by replacing the corresponding outcome models with the formulations above. In summary, the hotspot (mode of the $Y_t$ vector) is either Philadelphia ($L_t=1$) or Pittsburgh ($L_t=0$), where the probability depends on the covariates $\XB_t$. The values in the entries of $Y_t$ follow the log-likelihood of a $2$-dimensional isotropic Gaussian centered at the hotspot. As a result, the counterfactual and observed distributions will be bimodally distributed with different hotspot probabilities. We can then visually assess the performance of the models by comparing the distribution of the hotspot from the generated outcome samples to those of the counterfactual samples, as in Figure.~\ref{fig:semi-covid}.

\paragraph{Nationwide COVID-19 Mask data}

We perform a case study using real data by looking at the aggregated COVID-19 data sources from $2020$ to $2021$ spanning $49$ weeks due to the limited availability of the nationwide data.  We exclude $89$ counties with zero incremental new cases count. These counties either do not have a significant amount of infectious cases or have small populations, leading to $3,130$ counties across $56$ states/affiliated regions of the United States. For variables that only have state-level records, we map them to the county level for simplicity.

We analyze the same set of variables as the semi-synthetic COVID-19 dataset but exclude the vaccine dosage covariate because of missing data in some states. To align the outcome variable with the covariates and treatment, we set it to measure one week after these variables. Due to the long-tailed distribution of the outcome variable, we apply a base-$10$ logarithmic transformation during the modeling process.
Further details regarding the variables can be found in Table~\ref{append:tab-real}. 
We use the same model architecture described in Appendix~\ref{append:set-up} to train the IPTW network and the \texttt{MSCVAE}. We generate counterfactual outcomes for treatment combinations $\ab=(0,0,0)$ and $\ab=(1,1,1)$. Since other treatment combinations occur rarely (less than $5\%$ of observations), we exclude them from the final results.

\section{Additional synthetic results} \label{append:additional}
\subsection{Imbalanced treatment combinations}
\begin{figure*}[t!]
    \centering
     \includegraphics[width=0.8\linewidth]{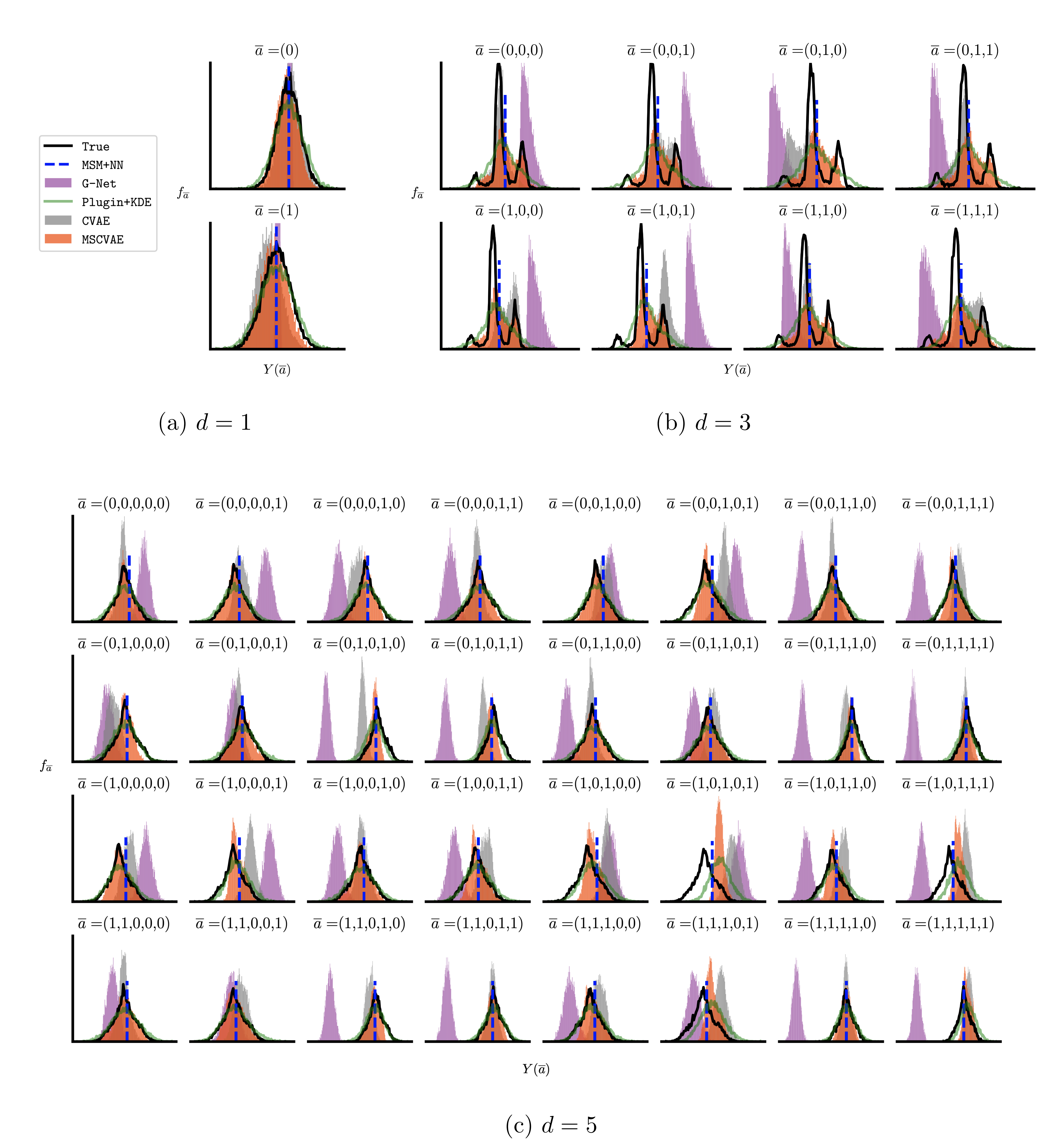}
\caption{\rebut{The estimated and true counterfactual distributions across various lengths of history dependence ($d=1,3,5)$ on synthetic datasets with imbalanced treatment. We exclude \texttt{KDE} for illustrative purposes.}}
\label{append:synthetic-imbalanced}
\end{figure*}

We provide additional fully-synthetic experiment while setting $\beta_0=-2$ (as opposed to $\beta_0=-0.5$,) where the treatment combinations are imbalancedly distributed.
Consistent with the findings in Figure~\ref{fig:synthetic-exps}, our results in Figure~\ref{append:synthetic-imbalanced} and Table~\ref{append:tab-imbalanced} demonstrate the superior performance of the \texttt{MSCVAE} model (represented by the orange shading) in accurately capturing the shape of the true counterfactual distributions (represented by the black line) across all scenarios. 
This observation further validates the quantitative comparisons presented in Table \ref{tab:comparision}, where \texttt{MSCVAE} achieves the smallest mean and Wasserstein distance among all baseline methods.
These results highlight that our algorithm attains competitive performance even when certain treatment combinations occur less frequently compared to others. 
This situation is common in real-life scenarios where certain treatment combinations are favored due to factors such as policy inertia.

\begin{table*}[t!]
 \caption {Quantitative performance on fully-synthetic data with imbalanced treatment combinations. }
 \centering
 \begin{adjustbox}{max width=1\textwidth}
 \begin{threeparttable}
 \begin{tabular}{ c   c   c  c c  c c  c  c} 
\toprule[1pt]\midrule[0.3pt]
\textbf{} &  \multicolumn{2}{c}{$d=1$} &  \multicolumn{2}{c}{$d=3$}& \multicolumn{2}{c}{$d=5$}  \\
\textbf{Methods} & {Mean $\downarrow$}& {Wasserstein $\downarrow$} &  {Mean $\downarrow$}& {Wasserstein $\downarrow$} & {Mean $\downarrow$}& {Wasserstein $\downarrow$}\\
\hline
\texttt{MSM+NN} & $\underline{0.009}$ ($\underline{0.013}$) &  $0.581$ ($0.583$) & $\textbf{0.069}$ ($\underline{0.633}$) &  $\textbf{0.145}$ ($0.657$)  & $0.173$ ($\textbf{0.448}$) &  $0.502$ ($\textbf{0.613}$)  \\

\texttt{KDE} & $0.253$ ($0.424$) &  $0.277$ ($0.424$) & $0.538$  ($1.313$)&  $0.573$ ($1.313$) & $0.518$ ($1.503$)  &  $0.520$ ($1.504$)  \\

\texttt{Plugin+KDE} & $\textbf{0.005}$ ($\textbf{0.006}$) &  $\textbf{0.122}$ ($\textbf{0.126}$) & $\underline{0.070}$ ($0.215$) &  $0.215$ ($\textbf{0.240}$)  & $\textbf{0.157}$ ($0.863$)  & $\underline{0.211}$ ($0.863$)  \\

\texttt{G-Net} & $0.068$ ($0.074$)  & $0.529$ ($0.532$) & $1.878$ ($3.004$) &  $1.881$ ($3.004$)  & $2.070$ ($4.794$)  & $2.072$ ($4.794$)  \\

\texttt{CVAE} & $0.254$ ($0.431$) &  $0.268$ ($0.431$) & $0.518$ ($1.244$) &  $0.569$ ($1.244$) & $0.521$ ($1.540$) &  $0.565$ ($1.540$)\\

\textcolor{orange}{\texttt{MSCVAE}} & $0.151$ ($0.152$) &  $\underline{0.152}$ ($\underline{0.153}$) & $0.175$ ($\textbf{0.332}$) &  $\underline{0.214}$  ($\underline{0.338}$) &$\underline{0.162}$ ($\underline{0.832}$) &  $\textbf{0.187}$ ($\underline{0.832}$)    \\

\midrule[0.3pt]\bottomrule[1pt]
\end{tabular}
\begin{tablenotes}[para,flushleft]
* Numbers represent the \textbf{average} metric across all treatment combinations and those in the parentheses represent the \textbf{worst} across treatment combinations. Bold and underlined numbers represent the best and second best results.
\end{tablenotes}
\end{threeparttable}
 \end{adjustbox}
  \label{append:tab-imbalanced}
\end{table*}

\newpage

\begin{table*}[t!]
 \caption {\rebut{Quantitative performance on fully-synthetic datasets with static baseline covariate}}
 \centering
 \begin{adjustbox}{max width=1\textwidth}
 \begin{threeparttable}
 \begin{tabular}{c c c c c c c c c c c c c} 
\toprule[1pt]\midrule[0.3pt]
\textbf{} & \multicolumn{6}{c}{\textbf{$V \in [-1,0]$}} & \multicolumn{6}{c}{\textbf{$V \in [0,1]$}} \\
\textbf{} &  \multicolumn{2}{c}{$d=1$} &  \multicolumn{2}{c}{$d=3$}& \multicolumn{2}{c}{$d=5$} & \multicolumn{2}{c}{$d=1$} &  \multicolumn{2}{c}{$d=3$}& \multicolumn{2}{c}{$d=5$}  \\
\textbf{Methods} & {Mean $\downarrow$}& {Wasserstein $\downarrow$} &  {Mean $\downarrow$}& {Wasserstein $\downarrow$} & {Mean $\downarrow$}& {Wasserstein $\downarrow$} &  {Mean $\downarrow$}& {Wasserstein $\downarrow$} &  {Mean $\downarrow$}& {Wasserstein $\downarrow$} & {Mean $\downarrow$}& {Wasserstein $\downarrow$} \\
\hline
\texttt{MSM+NN} & $\textbf{0.002}$ ($\textbf{0.003}$) &  $0.408$ ($0.408$) & $\underline{0.064}$ ($\textbf{0.104}$) &  $0.449$ ($0.466$)  & $\textbf{0.164}$ ($\textbf{0.441}$) &  $0.368$ ($\textbf{0.517}$) & $0.015$ ($0.019$) &  $0.407$ ($0.412$)&  $0.057$ ($\textbf{0.068}$) &  $0.466$ ($0.475$) &  $0.182$ ($\textbf{0.484}$) &  $0.388$ ($\textbf{0.549}$) \\

\texttt{KDE}  & $0.194$ ($0.265$) &  $0.201$ ($0.266$) & $0.557$ ($1.261$) &  $0.562$ ($1.261$)  & $0.562$ ($1.590$) &  $0.564$ ($1.590$) & $0.216$ ($0.251$) &  $0.222$ ($0.256$)&  $0.547$ ($1.138$) &  $0.548$ ($1.138$) &  $0.559$ ($1.517$) &  $0.561$ ($1.517$) \\

\texttt{Plugin+KDE} & $0.010$ ($0.013$) &  $\underline{0.117}$ ($\underline{0.121}$) & $0.073$ ($0.230$) &  $\underline{0.134}$ ($\underline{0.230}$)   &  $0.170$ ($0.823$) & $\underline{0.196}$ ($0.823$) &  $\underline{0.009}$ ($\underline{0.017}$)&  $\underline{0.121}$ ($\underline{0.130}$) &  $\underline{0.055}$ ($0.165$) &  $\underline{0.109}$ ($\underline{0.165}$) &  $\textbf{0.148}$ ($0.670$) & $\underline{0.182}$ ($0.670$) \\

\texttt{G-Net}  & $0.431$ ($0.512$) &  $0.431$ ($0.512$) & $0.819$ ($1.885$) &  $0.823$ ($1.885$)  & $0.815$ ($2.238$) &  $0.843$ ($2.238$) & $0.452$ ($0.551$) &  $0.452$ ($0.551$)&  $0.738$ ($1.707$) &  $0.739$ ($1.707$) &  $0.735$ ($2.076$) &  $0.768$ ($2.076$) \\

\texttt{CVAE}  & $0.238$ ($0.327$) &  $0.240$ ($0.328$) & $0.537$ ($1.192$) &  $0.571$ ($1.192$)  & $0.534$ ($1.478$) &  $0.585$ ($1.478$) & $0.256$ ($0.290$) &  $0.258$ ($0.290$)&  $0.534$ ($1.094$) &  $0.565$ ($1.094$) &  $0.534$ ($1.423$) &  $0.591$ ($1.423$) \\

\textcolor{orange}{\texttt{MSCVAE}}  & $\underline{0.009}$ ($\underline{0.010}$) &  $\textbf{0.051}$ ($\textbf{0.056}$) & $\textbf{0.058}$ ($\underline{0.181}$) &  $\textbf{0.083}$ ($\textbf{0.188}$)  & $\underline{0.169}$ ($\underline{0.767}$) &  $\textbf{0.186}$ ($\underline{0.767}$) & $\textbf{0.006}$ ($\textbf{0.010}$) &  $\textbf{0.047}$ ($\textbf{0.055}$)&  $\textbf{0.046}$ ($\underline{0.140}$) &  $\textbf{0.068}$ ($\textbf{0.158}$) &  $\underline{0.151}$ ($\underline{0.659}$) &  $\textbf{0.172}$ ($\underline{0.659}$) \\
\midrule[0.3pt]\bottomrule[1pt]
\end{tabular}
\begin{tablenotes}[para,flushleft]
* Numbers represent the \textbf{average} metric across all treatment combinations and those in the parentheses represent the \textbf{worst} across treatment combinations. Bold and underlined numbers represent the best and second best results.
\end{tablenotes}
\end{threeparttable}
 \end{adjustbox}
 \label{append:cate}
\end{table*}

\begin{figure*}[t!]
    \centering
     \includegraphics[width=0.8\linewidth]{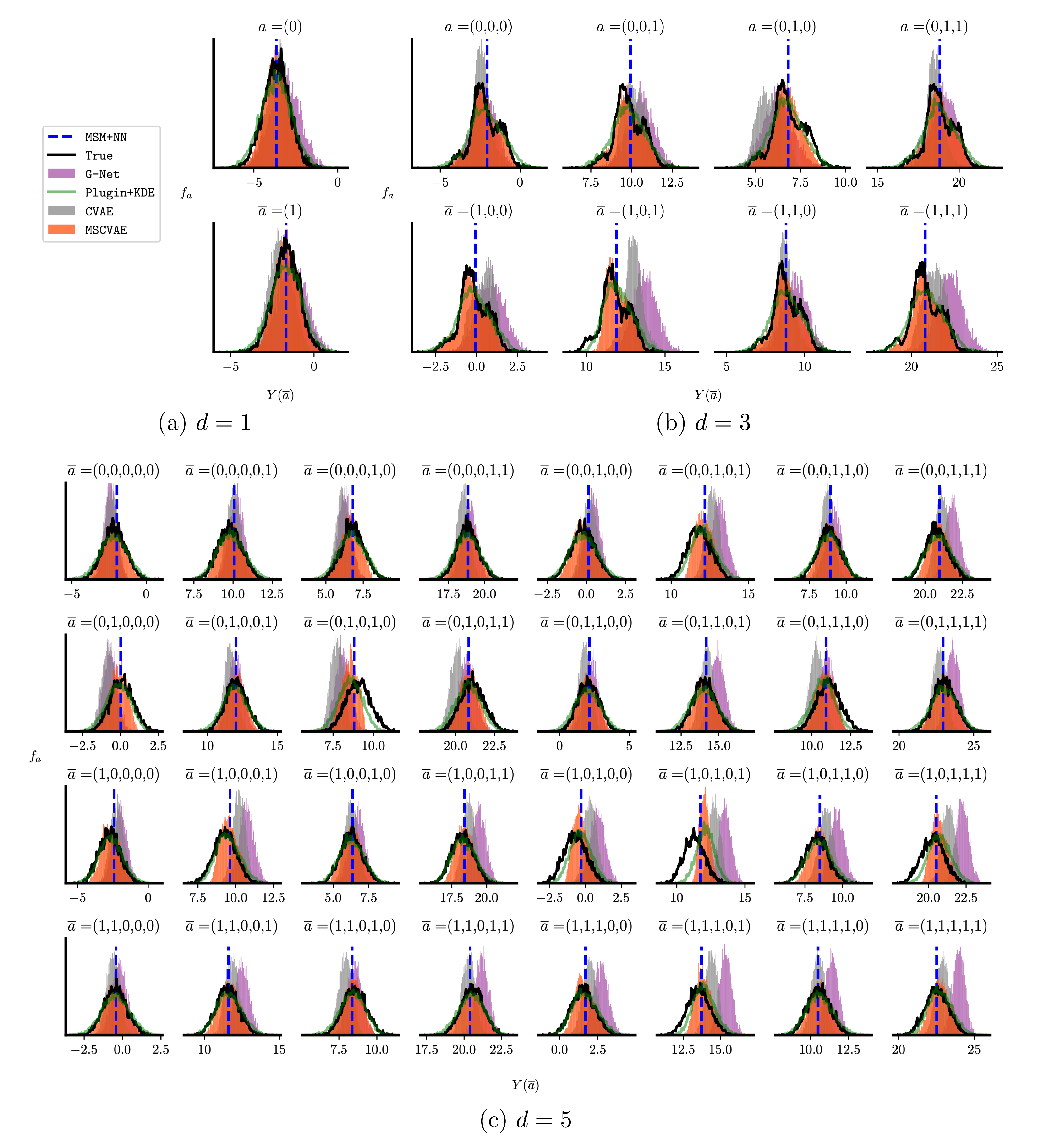}
\caption{\rebut{The estimated and true counterfactual distributions across various lengths of history dependence ($d=1,3,5)$ on synthetic datasets with $V\in [-1,0]$. Each sub-panel provides a comparison for a specific treatment combination $\ab$. We exclude \texttt{KDE} for illustrative purposes.}}
\label{append:synthetic-exps-cate0}
\end{figure*}
\newpage

\begin{figure*}[t!]
    \centering
     \includegraphics[width=0.8\linewidth]{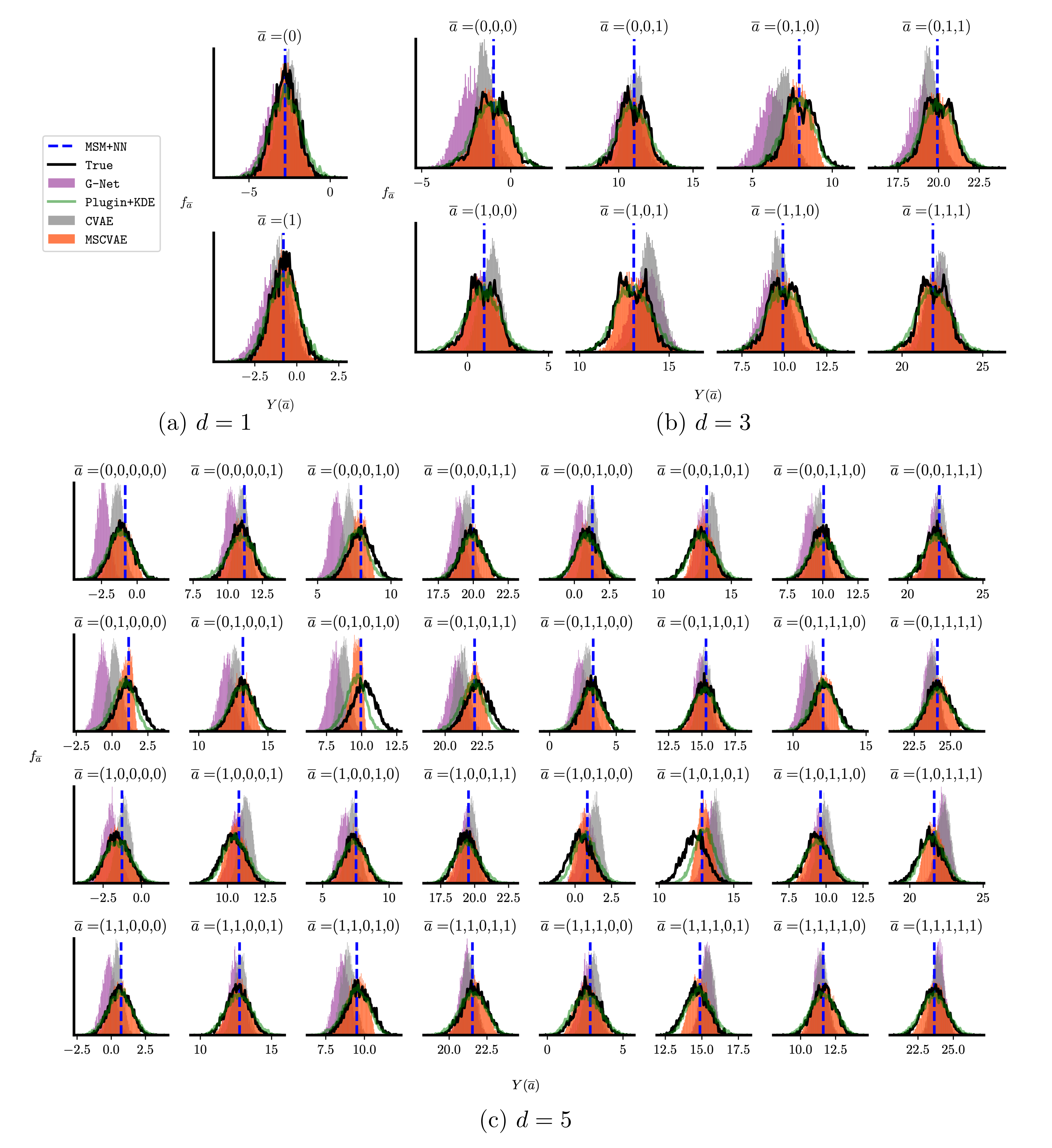}

\caption{\rebut{The estimated and true counterfactual distributions across various lengths of history dependence ($d=1,3,5)$ on synthetic datasets with $V\in [0,1]$. Each sub-panel provides a comparison for a specific treatment combination $\ab$. We exclude \texttt{KDE} for illustrative purposes.}}
\label{append:synthetic-exps-cate1}
\end{figure*}

\newpage
\subsection{Generating samples from conditional counterfactual distribution}

 A related extension of our work might be to infer conditional counterfactual outcomes (related to the conditional average treatment effect, CATE). This corresponds to looking at the distribution of the outcome under a specific subpopulation. Since our covariates are assumed to be time-varying, a common approach is to introduce a set of static baseline covariates, $V$ \citep{robins1999estimation}. The baseline covariate $V$ denotes the static feature (such as a patient's gender or age) that will influence both the time-varying covariates $X$ and the outcome $Y$. We can then draw counterfactual samples from a specific sub-population by conditioning on the values of the $V$. In this framework, we observe ($y_t^i$, $
 \ab_t^i$, $\xb_t^i$, $v^i$), where $v \in \mathbb{R}^{\nu}$ is a static baseline variable tha differs by individual. Accordingly, the generator will have an additional input   $g_\theta(z, \ab,v): \mathbb{R}^r \times \mathcal{A}^d \times \mathbb{R}^{\nu} \rightarrow \mathcal{Y}$, and the IPTW weights will become $ w_\phi(\ab,\xb,v) = \frac{1}{\prod_{\tau=t-d+1}^{t} f_\phi(a_{\tau}|\ab_{\tau-1},\xb_{\tau},v)}$. Correspondingly, the objective in Proposition~\ref{prop:pesudo-objective} will become: 
\[\EE_{v} ~\EE_{\AB} ~\left[\EE_{y \sim f_{\ab,v}} ~\log f_\theta(y|\ab,v)\right] \approx
    \frac{1}{N}\sum_{(y,\ab,\xb,v) \in \mathcal{D}}w_\phi(\ab,\xb,v) \log f_\theta(y|\ab,v).
    \]
We have conducted additional experiments for the fully synthetic data, where the baseline variable, $V$, was divided into two groups. The $V$ was uniformly drawn from $[-1,0]$ in the first group and from $[0,1]$ in the second group. We then looked at the performance of the generative samples corresponding to two sub-groups generating samples with the corresponding treatment combinations and the baseline covariate, $V$. Specifically, for the \texttt{MSCVAE} and \texttt{CVAE}, under a specific $\ab$, we generated $v$ from $[-1,0]$ and $[0,1]$  uniformly at random and fed them into the conditional generators, along with the Gaussian noises, $z$.  We have included the results in Table~\ref{append:cate},  Fig.~\ref{append:synthetic-exps-cate0} ($V\in[-1,0]$), and Fig.~\ref{append:synthetic-exps-cate1} ($V\in[0,1]$). It can be seen that \textit{MSCVAE} still outperforms other baselines.

\section{Code availability}
The implementation code is available at \url{https://github.com/ShenghaoWu/Counterfactual-Generative-Models}.

\end{document}